\documentclass[10pt,journal,compsoc]{IEEEtran}
\usepackage{amsmath,amsfonts}
\usepackage{algorithmic}
\usepackage{algorithm}
\usepackage{array}
\usepackage[caption=false,font=normalsize,labelfont=sf,textfont=sf]{subfig}
\usepackage{textcomp}
\usepackage{stfloats}
\usepackage{url}
\usepackage{verbatim}
\usepackage{graphicx}
\usepackage{times}
\usepackage{epsfig}
\usepackage{amsmath}
\usepackage{amssymb}
\usepackage{paralist}

\usepackage{booktabs}
\usepackage{multicol}
\usepackage{multirow}
\usepackage{bm}
\usepackage{bbm}
\usepackage{diagbox}
\usepackage{makecell}
\usepackage{color, colortbl}
\usepackage[backref]{hyperref}
\usepackage{marvosym}
\definecolor{green}{rgb}{0.01, 0.75, 0.24}

\newcommand{\gr}{\rowcolor[gray]{.95}} 

\def\ours{RCBEVDet}

\renewcommand{\eqref}[1]{Eq.~(\ref{#1})}

\ifCLASSOPTIONcompsoc
  \usepackage[nocompress]{cite}
\else
  \usepackage{cite}
\fi
\ifCLASSINFOpdf
\else
\fi

\begin{document}
%
\title{RCBEVDet++: Toward High-accuracy Radar-Camera Fusion 3D Perception Network}
%
%
%
%

\author{Zhiwei Lin$^{*}$, 
        Zhe Liu$^{*}$, 
        Yongtao Wang\textsuperscript{\Letter}, 
        Le Zhang,
        and Ce Zhu
\IEEEcompsocitemizethanks{
\IEEEcompsocthanksitem Corresponding author: Yongtao Wang. $^{*}$ indicates equal contribution.
\IEEEcompsocthanksitem Zhiwei~Lin and Yongtao~Wang are with Wangxuan Institute of Computer Technology, Peking University, Beijing 100080, China.
E-mail: \{zwlin, wyt\}@pku.edu.cn
\IEEEcompsocthanksitem Zhe~Liu, Le~Zhang, and Ce~Zhu are with School of Information and Communication Engineering, University of Electronic Science and Technology of China, Sichuan 611731, China.
E-mail: liuzhe@std.uestc.edu.cn, \{lezhang,eczhu\}@uestc.edu.cn
\IEEEcompsocthanksitem A preliminary version of this manuscript was published in \cite{lin2024rcbevdet}
}
}

\IEEEtitleabstractindextext{%
\begin{abstract}
Perceiving the surrounding environment is a fundamental task in autonomous driving. To obtain highly accurate and robust perception results, modern autonomous driving systems typically employ multi-modal sensors, such as LiDAR, multi-view cameras, and millimeter-wave radar, to collect comprehensive environmental data. Among these, the radar-camera multi-modal perception system is especially favored for its excellent sensing capabilities and cost-effectiveness. However, the substantial modality differences between millimeter-wave radar and multi-view camera sensors pose significant challenges in fusing information from these two types of sensors. To address this problem, this paper presents RCBEVDet, a radar-camera fusion 3D object detection framework. Specifically, RCBEVDet is developed from an existing camera-based 3D object detection model, supplemented by a specially designed radar feature extractor, RadarBEVNet, and a radar-camera Cross-Attention Multi-layer Fusion (CAMF) module. Firstly, RadarBEVNet encodes sparse radar points into a dense bird’s-eye-view (BEV) feature using a dual-stream radar backbone and a Radar Cross Section (RCS) aware BEV encoder. Secondly, the CAMF module utilizes a deformable attention mechanism to align radar and camera BEV features and adopts channel and spatial fusion layers to fuse these multi-modal features. To further enhance RCBEVDet’s capabilities, we introduce RCBEVDet++, which advances the CAMF through sparse fusion, supports query-based multi-view camera perception models, and adapts to a broader range of perception tasks. Extensive experiments on the nuScenes dataset demonstrate that our method integrates seamlessly with existing camera-based 3D perception models and improves their performance across various perception tasks. Furthermore, our method achieves state-of-the-art radar-camera fusion results in 3D object detection, BEV semantic segmentation, and 3D multi-object tracking tasks. Notably, with ViT-L as the image backbone, RCBEVDet++ achieves 72.73 NDS and 67.34 mAP in 3D object detection without test-time augmentation or model ensembling. The source code and models will be released at \url{https://github.com/VDIGPKU/RCBEVDet}.
\end{abstract}

}

\maketitle

\IEEEdisplaynontitleabstractindextext

%
\IEEEpeerreviewmaketitle

\IEEEraisesectionheading{
\section{Introduction}\label{sec:introduction}}
\IEEEPARstart{A}{utonomous}
driving aims to improve safety, efficiency, and convenience in transportation by developing systems that allow vehicles to operate without human intervention~\cite{ma20233dsurvey, grigorescu2020survey}.
A major challenge for these systems is to perceive the surrounding environment as comprehensively as humans do, which is crucial for accurate trajectory prediction and motion planning.
To achieve this, modern autonomous driving systems primarily employ three types of sensors, \textit{e.g.}, multi-view cameras, millimeter-wave radar, and LiDAR, to gather information about the surrounding environment.

Among these types of sensors, the LiDAR sensor provides detailed geometric information that significantly enhances the perception process, resulting in optimal performance~\cite{DBLP:journals/corr/bevfusionmit}.
However, high-quality LiDAR sensors are expensive, which increases manufacturing costs. 
In contrast, multi-view cameras and millimeter-wave radar sensors offer more economical alternatives for both manufacturers and users. 
Compared with LiDAR, multi-view cameras capture intricate details such as color and texture, offering high-resolution semantic information, while the millimeter-wave radar sensor is superior in distance and velocity estimation~\cite{charvat2014small} and performs reliably under diverse weather and lighting conditions~\cite{zhou2022towards,li2022exploiting}.
Besides, advancements in 4D millimeter-wave radar technology are gradually overcoming its limitation of sparse radar points, positioning it as a potential substitute~\cite{bijelic2020seeing}.
Despite these advantages, a notable performance gap remains between LiDAR and camera or radar-based perception models.
A practical and effective strategy to bridge this gap involves integrating millimeter-wave radar with multi-view cameras, which complement each other and result in a more comprehensive and reliable perception.

To fuse radar and image data, recent works~\cite{zheng2023rcfusion,xiong2023lxl} primarily adopt the BEVFusion pipeline~\cite{DBLP:journals/corr/bevfusion, DBLP:journals/corr/bevfusionmit} by projecting both multi-view image features and radar features into a bird's eye view (BEV). 
However, simple fusion techniques such as concatenation or summation, as employed by BEVFusion, fail to address spatial misalignment between multi-view images and radar inputs.
Moreover, most radar-camera fusion methods~\cite{kim2023craft, Kim_2023_ICCVCRN,nabati2021centerfusion} still utilize encoders originally designed for LiDAR points, such as PointPillars, to extract radar features. 
Though these methods yield commendable results, the LiDAR-specific encoders they use do not account for unique radar characteristics, such as the Radar Cross Section (RCS), resulting in sub-optimal performance.

In this paper, we introduce RCBEVDet, a novel framework that effectively fuses radar and camera features in BEV space for the 3D object detection task. 
To address the unique characteristics of radar inputs, we specially design RadarBEVNet for efficient radar BEV feature extraction.
Specifically, RadarBEVNet begins by encoding radar inputs into distinct point-based and transformer-based representations through a dual-stream radar encoder.
Besides, an Injection and Extraction module is implemented to integrate features from these two representations.
Subsequently, these features are transformed into BEV features via RCS-aware scattering, which leverages RCS as a prior for object size and allocates point features throughout the BEV space.
In addition to RadarBEVNet, RCBEVDet integrates a Cross-Attention Multi-layer Fusion Module (CAMF) to achieve a robust fusion of radar and camera features within the BEV space. 
To be more specific, CAMF employs multi-modal cross-attention to adaptively correct coordinate mismatches between the two types of BEV features. 
Then, channel and spatial fusion layers are applied to further consolidate the multi-modal features, enhancing the overall detection performance.

To fully leverage the capabilities of RCBEVDet, we have upgraded the CAMF module with sparse fusion to support query-based multi-view camera perception models. 
Additionally, we have broadened RCBEVDet's functionality to encompass a broader range of
perception tasks, including 3D object detection, BEV semantic segmentation, and 3D multi-object tracking. 
This enhanced framework is named RCBEVDet++. 
Specifically, to adapt the CAMF module for query-based multi-view camera methods, which lack explicit camera BEV features, we replace the original camera BEV features with camera object queries that incorporate 3D coordinates.
This modification develops a new query component in our multi-modal cross-attention mechanism. 
Then, we perform a project-and-sample process, where camera object queries are projected into the BEV space and matched with corresponding radar features to form radar object queries. 
Subsequently, the multi-modal queries are aligned using deformable cross-attention. 
Finally, the adjusted multi-modal queries are concatenated and sent to a simple linear layer for effective feature fusion, boosting perception performance across the expanded range of tasks.

The main contributions of this paper are listed as follows:
\begin{itemize}
    \item We introduce RCBEVDet, a radar-camera fusion framework designed for highly accurate and robust 3D object detection, which consists of RadarBEVNet for radar BEV feature extraction and the Cross-Attention Multi-layer Fusion Module (CAMF) for robust radar-camera feature fusion in BEV space.
    \item 
    Based on RCBEVDet, we further propose RCBEVDet++ perception framework, which extends the CAMF module to accommodate query-based multi-view camera perception models and unleashes the full potential of RCBEVDet in various 3D perception tasks.
    \item 
   On the nuScenes benchmark, RCBEVDet improves the performance of camera-based 3D object detectors and demonstrates robust capabilities against sensor failure cases. Additionally, RCBEVDet++ further enhances camera-based perception models, achieving state-of-the-art results in radar-camera multi-modal 3D object detection, BEV semantic segmentation, and 3D multi-object tracking tasks.
\end{itemize}

\section{Related Work}
\subsection{Camera-based 3D Perception}
3D object detection, BEV semantic segmentation, and 3D multi-object tracking are three fundamental perception tasks for autonomous driving. 
Currently, many 3D multi-object tracking methods often adopt the Tracking-By-Detection framework, which utilizes results from 3D object detection to associate objects.
These tracking methods focus on object matching rather than efficiently processing input images.
Besides, more accurate detection results can bring higher tracking performance.
Therefore, in this section, we mainly discuss more diverse 3D object detection and BEV semantic segmentation methods that process multi-frame multi-view camera inputs.
Specifically, 3D object detection aims to predict the location and category of 3D objects, while semantic segmentation integrates vehicle recognition, semantic lane map prediction, and drivable area estimation tasks. 
However, detecting objects and segmenting maps in 3D space using camera images is challenging due to insufficient 3D information.
In recent years, numerous studies \cite{Graph-DETR3D, huang2021bevdet, li2022bevformer, shi2020points, weng2019monocular, chen2020monopair} have made substantial efforts to address this issue, including inferring depth from images \cite{xu2021monocular}, utilizing geometric constraints and shape priors \cite{lu2021geometry}, designing specific loss functions \cite{simonelli2020towards, miao2021pvgnet}, and exploring joint 3D detection and reconstruction optimization \cite{liu2021voxel}. 
More recently, multi-view sensors have become a popular configuration for autonomous vehicles, providing more comprehensive surrounding information. The emergence of multi-view camera datasets \cite{nuscenes, sun2020scalabilityWaymo} has led to the development of multi-view 3D object detection and BEV semantic segmentation methods \cite{philion2020liftLSS, huang2021bevdet, li2023bevdepth, huang2022bevdet4d, wang2021detr3d, li2022bevformer, liu2022petr, Liu_2023_ICCVSparseBEV, wang2023exploringStreamPETR, Zhou_2022_CVPRcvt, Man_2023_CVPRBEVguided, schramm2024bevcar}, which can be broadly categorized into two approaches: geometry-based methods and transformer-based methods.

\subsubsection{Geometry-based Methods}
Geometry-based multi-view 3D object detection and BEV semantic segmentation predominantly utilize depth prediction networks to estimate the depth distribution in images. 
This facilitates the conversion of extracted 2D image features into 3D camera frustum features. 
Subsequently, operations such as Voxel Pooling are employed to construct features in the voxel or BEV space.

Specifically, as a pioneering work, Lift-Splat-Shoot (LSS)~\cite{philion2020liftLSS} first employs a lightweight depth prediction network to explicitly estimate the depth distribution and a context vector for each image. 
Then, the outer product of the depth and context vector determines the feature at each point in 3D space along the perspective ray, enabling the effective transformation of image features into BEV features.
Building on LSS, FIERY \cite{fiery2021} introduces a future instance prediction model based on BEV, capable of predicting the future instances of dynamic agents and their motions. 
BEVDet \cite{huang2021bevdet} extends the viewpoint transformation technique from LSS to detect 3D objects using BEV features. 
Additionally, BEVDepth \cite{li2023bevdepth} leverages explicit depth information from LiDAR as supervision to enhance depth estimation and incorporates camera extrinsic parameters as a prior for depth estimation.
Based on BEVDet, BEVDet4D \cite{huang2022bevdet4d} performs spatial alignment of BEV features across historical frames, significantly improving detection performance. 
Furthermore, SOLOFusion \cite{DBLP:conf/iclr/solofusion} proposes to fuse high-resolution short-term and low-resolution long-term features, enhancing the inference speed of 3D detection with long-term temporal inputs.

\subsubsection{Transformer-based Methods}
Transformer-based methods leverage attention mechanisms to project predefined queries onto multi-view image planes using coordinate transformation matrices, subsequently updating the query features with multi-view image features.
Specifically, the pioneering work DETR3D \cite{wang2021detr3d} employs a transformer decoder for 3D object detection, developing a top-down framework and utilizing a set-to-set loss to measure the difference between ground truth and predictions. 
Similarly, CVT \cite{Zhou_2022_CVPRcvt} introduces a simple BEV semantic segmentation baseline using a cross-view transformer architecture. Following this, BEVformer \cite{li2022bevformer} constructs dense BEV queries and incorporates multi-scale deformable attention to map multi-view image features to these dense queries. 
Moreover, PETR \cite{liu2022petr} generates multi-view image features with explicit position information derived from 3D coordinates. 
Building on PETR, PETRv2 \cite{liu2023petrv2} integrates temporal fusion across multiple frames and extends 3D positional embedding with time-aware modeling. 
Additionally, Sparse4D \cite{Sparse4D} allocates and projects multiple 4D key points for each 3D anchor to generate different views, aspect ratio, and timestamp features, which are then fused hierarchically to improve the overall image feature representation. Sparse4Dv2 \cite{Sparse4Dv2} extends Sparse4D with a more efficient time fusion module and introduces camera parameter encoding and dense depth supervision.
More recently, StreamPETR \cite{wang2023exploringStreamPETR} utilizes sparse object queries as intermediate representations to capture temporal information, and SparseBEV \cite{Liu_2023_ICCVSparseBEV} incorporates a scale-adaptive self-attention module and an adaptive spatio-temporal sampling module to dynamically capture BEV and temporal information.

\begin{figure*}[!t]
    \centering
    \includegraphics[width=0.95\linewidth]{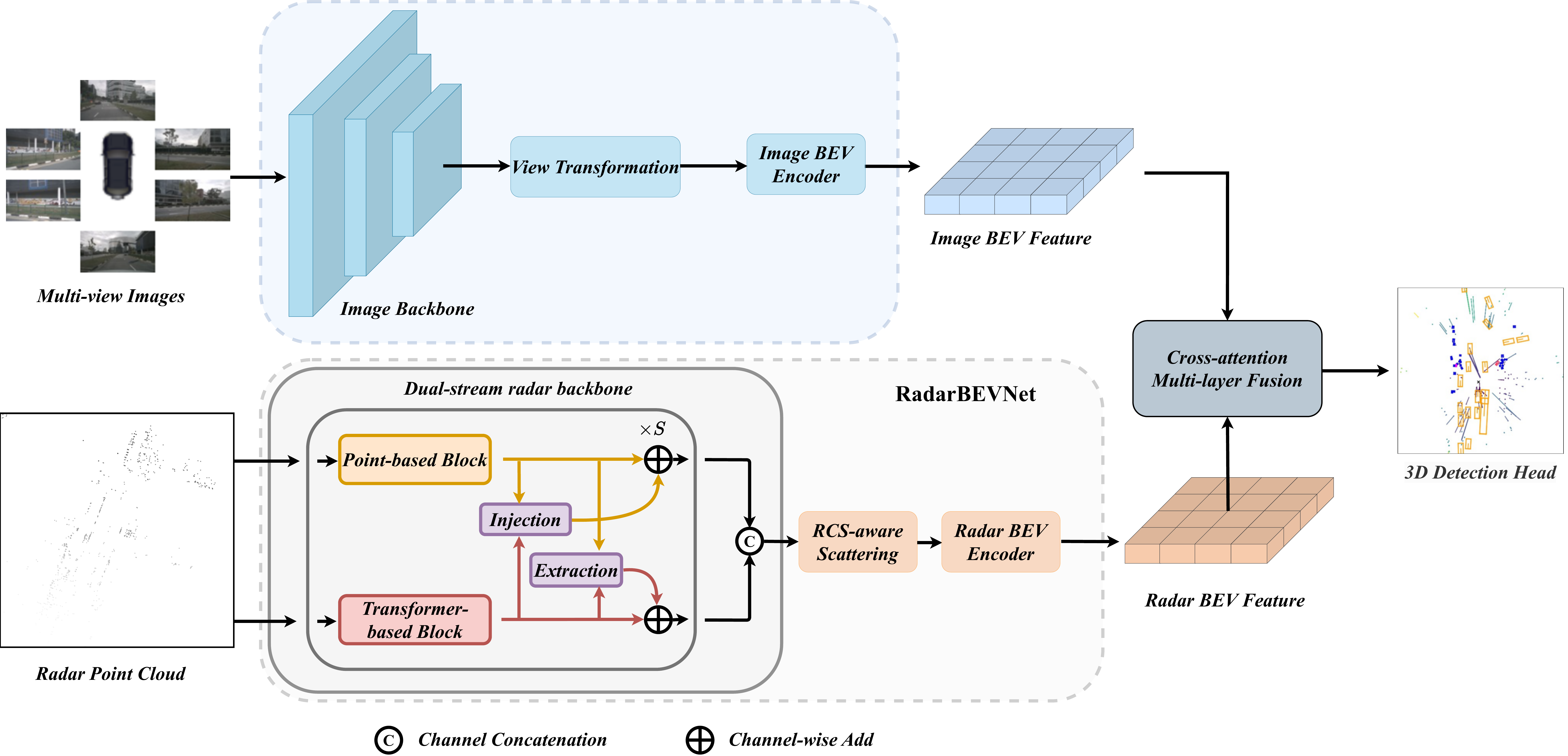}
    \caption{\textbf{Overall pipeline of \ours.}
    Firstly, multi-view images are encoded and transformed into image BEV features. Concurrently, radar point clouds are processed by the proposed RadarBEVNet to extract radar BEV features. Subsequently, features from both radar and cameras are dynamically aligned and aggregated using the Cross-Attention Multi-layer Fusion (CAMF) module. The resulting semantically rich multi-modal feature is then utilized for the 3D object detection task.
    }
    \label{fig:pipeline}
\end{figure*}

\subsection{Radar-camera 3D Perception}
Millimeter-wave radar is a widely used sensor in autonomous vehicles due to its cost-effectiveness, long-range perception, Doppler velocity measurement, and robustness against adverse weather conditions. 
Although millimeter-wave radar data generally includes distance, angle, and velocity information, it performs relatively poorly in measuring targets' pitch angles. 
Additionally, the inherent sparsity and lack of semantic information in millimeter-wave radar data pose challenges for pure radar-based 3D perception. 
As a result, millimeter-wave radar is frequently employed as an auxiliary modality to enhance the performance of multi-modal 3D perception systems.

In recent years, the combination of multi-view cameras and millimeter-wave radar sensors for 3D perception has attracted significant attention, owing to the complementary nature of the information provided by these two modalities.
Concretely, RadarNet \cite{yang2020radarnet} introduces a multi-level radar-camera fusion pipeline to improve the accuracy of distant object detection and reduce velocity errors. 
CenterFusion \cite{nabati2021centerfusion} utilizes a keypoint detection network to generate initial 3D detection results from images and incorporates a pillar-based radar association module to refine these results by linking radar features with corresponding detected boxes. 
Similarly, MVFusion \cite{DBLP:conf/icra/mvfusion} achieves semantic alignment between cameras and millimeter-wave radar, enhancing the interaction between the two modalities. 
Additionally, Simple-BEV \cite{Harley2023Simple-BEV} investigates architecture designs and hyper-parameter settings for multi-sensor BEV perception systems. 
CRAFT \cite{kim2023craft} proposes a proposal-level fusion framework that employs a Soft-Polar-Association and Spatio-Contextual Fusion Transformer to efficiently exchange information between the camera and millimeter-wave radar. 
RADIANT \cite{long2023radiant} develops a network to estimate positional offsets between radar echoes and object centers and leverages radar depth information to enhance camera features. 
More recently, CRN \cite{Kim_2023_ICCVCRN} generates radar-augmented image features with radar depth information for multi-view transformation and incorporates a cross-attention mechanism to address spatial misalignment and information disparity between radar and camera sensors. 
RCFusion \cite{zheng2023rcfusion} utilizes radar PillarNet~\cite{shi2022pillarnet} to generate radar pseudo-images and presents a weighted fusion module to effectively fuse radar and camera BEV features.
BEVGuide~\cite{Man_2023_CVPRBEVguided} builds on the CVT~\cite{Zhou_2022_CVPRcvt} framework and proposes a sensor-agnostic attention module based on BEV, facilitating BEV representations learning and understanding.
BEVCar~\cite{schramm2024bevcar} introduces an innovative radar-camera fusion method for BEV map and object segmentation using an attention-based image lift strategy.

\section{RCBEVDet: Radar-camera fusion in BEV for 3D object detection}
\label{sec:rcbevdet}
The overall pipeline of RCBEVDet is illustrated in Figure \ref {fig:pipeline}. Specifically, multi-view images are processed by an image encoder to extract features, which are then transformed into the image BEV feature using a view-transformation module. Concurrently, radar point clouds are encoded into the radar BEV feature by the proposed RadarBEVNet. Next, the image and radar BEV features are fused using the Cross-Attention Multi-layer Fusion module. Finally, the fused multi-modal BEV feature is employed for the 3D object detection task.

\subsection{RadarBEVNet}
\label{Sec:dual-stream encoder}
Previous radar-camera fusion methods typically utilize radar encoders designed for LiDAR point clouds, such as PointPillars \cite{lang2019pointpillars}. In contrast, we introduce RadarBEVNet, specifically tailored for efficient radar BEV feature extraction. 
RadarBEVNet encodes sparse radar points into a dense BEV feature using a dual-stream radar backbone and an RCS-aware BEV encoder. More specifically, the dual-stream radar backbone processes radar points into two representations: a local point-based representation and a global transformer-based representation, employing an Injection and Extraction module to fuse these representations. The RCS-aware BEV encoder leverages RCS as an object size prior, distributing the feature of a single radar point across multiple pixels in the BEV space.

\subsubsection{Dual-stream radar backbone}

The dual-stream radar backbone consists of two components: a point-based backbone and a transformer-based backbone. The point-based backbone focuses on learning local radar features, while the transformer-based backbone captures global information.

\begin{figure}[!t]
\centering
\subfloat[]{\includegraphics[width=0.48\linewidth]{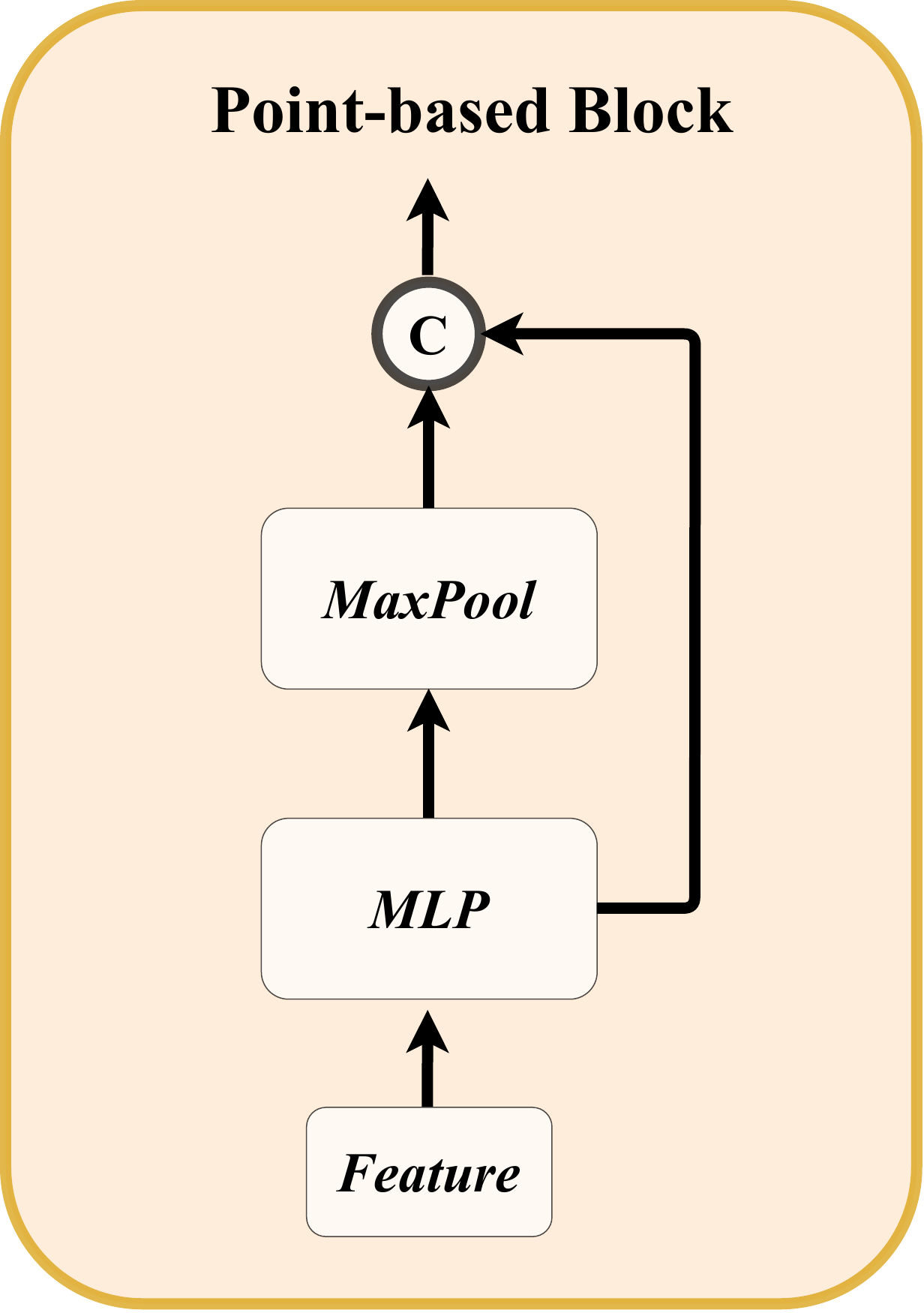}%
\label{fig:point block}}
\hfil
\subfloat[]{\includegraphics[width=0.48\linewidth]{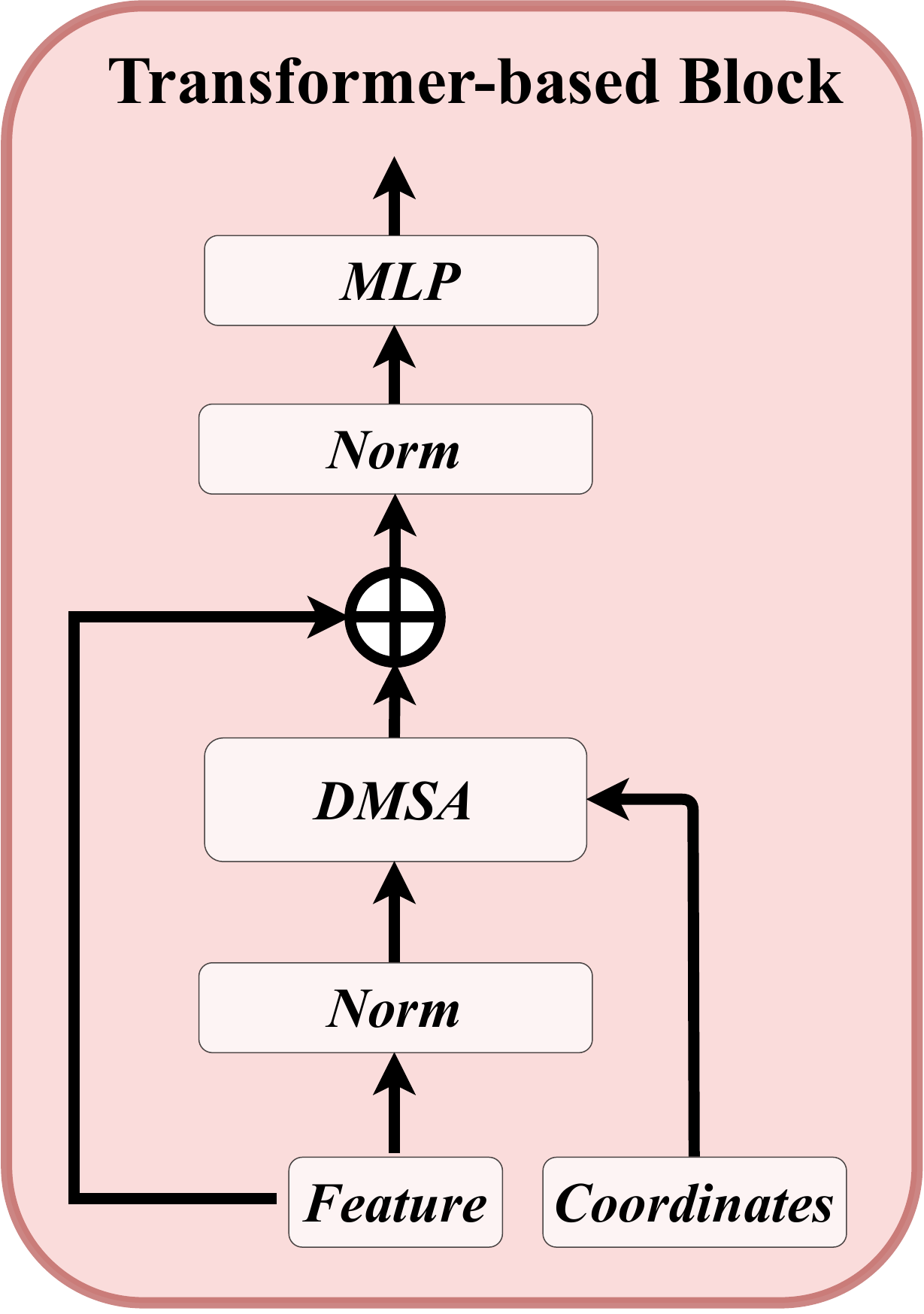}%
\label{fig:transformer block}}
\caption{\textbf{Architecture of the dual-stream radar backbone.} (a) Point-based Block. (b) Transformer-based Block.}
\label{fig:dual stream backbone}
\end{figure}

For the point-based backbone, we adopt an architecture similar to PointNet \cite{qi2017pointnet}. As illustrated in Figure \ref{fig:point block}, the point-based backbone comprises $S$ blocks, each containing a multi-layer perceptron (MLP) and a max pooling operation. Specifically, the input radar point feature $f$ is first processed by the MLP to increase its feature dimension. 
Then, the high-dimensional radar features are sent to the MaxPool layer with a residual connection.
The whole process can be formulated as follows:
\begin{equation}
    f = \text{Concat}[\text{MLP}(f),\text{MaxPool}(\text{MLP}(f))].
\end{equation}

As for the transformer-based backbone, it comprises $S$ standard transformer blocks \cite{vaswani2017attention, dosovitskiy2020vit}, which includes an attention mechanism, a feed-forward network, and normalization layers, as shown in Figure \ref{fig:transformer block}. Due to the extensive range of autonomous driving scenarios, optimizing the model directly using standard self-attention can be challenging. To address this issue, we propose a distance-modulated self-attention mechanism (DMSA) to facilitate model convergence by aggregating neighbor information during the early training iterations. More specifically, given the coordinates of $N$ radar points, we first calculate the pair-distance $D\in \mathbb{R}^{N\times N}$ between all points. 
Then, we generate the Gaussian-like weight map $G$ according to the pair-distance $D$ as follows:
\begin{equation}
    G_{i,j} = \exp(-D_{i,j}^2/\sigma^2),
\end{equation}
where $\sigma$ is a learnable parameter to control the bandwidth of the Gaussian-like distribution.
Essentially, the Gaussian-like weight map $G$ assigns high weight to spatial locations near the point and low weight to positions far from the point.
We can modulate the attention mechanism with the generated weight $G$ as follows:
\begin{equation}
\begin{split}
    \text{DMSA}(Q, K, V) = &\text{Softmax}(\frac{QK^{\top}}{\sqrt{d}}+\log G)V \\
    =&\text{Softmax}(\frac{QK^{\top}}{\sqrt{d}}-\frac{1}{\sigma^2}D^2)V,
\end{split}
\end{equation}
where $Q$, $K$, and $V$ denote query, key, and value in attention mechanism~\cite{vaswani2017attention}.
To ensure DMSA can be degraded to vanilla self-attention, we replace $1/\sigma$ with a trainable parameter $\beta$ during the training.
When $\beta~\text{=}~0$, DMSA is degraded to the vanilla self-attention.
We also investigate the multi-head DMSA. Each head has unshared $\beta_i$ to control the receptive field of DMSA.
The multi-head DMSA with $H$ heads can be formulated as $\text{MultiHeadDMSA}(Q, K, V)=\text{Concat}[head_1, head_2, ..., head_H ]$, where
\begin{equation}
\begin{split}
    head_i &= DMSA(Q_i, K_i, V_i) \\
    &=\text{Softmax}(\frac{Q_iK_i^{\top}}{\sqrt{d_i}}-\beta_iD^2)V_i. \\
\end{split}
\end{equation}

\begin{figure}
    \centering
    \includegraphics[width=0.95\linewidth]{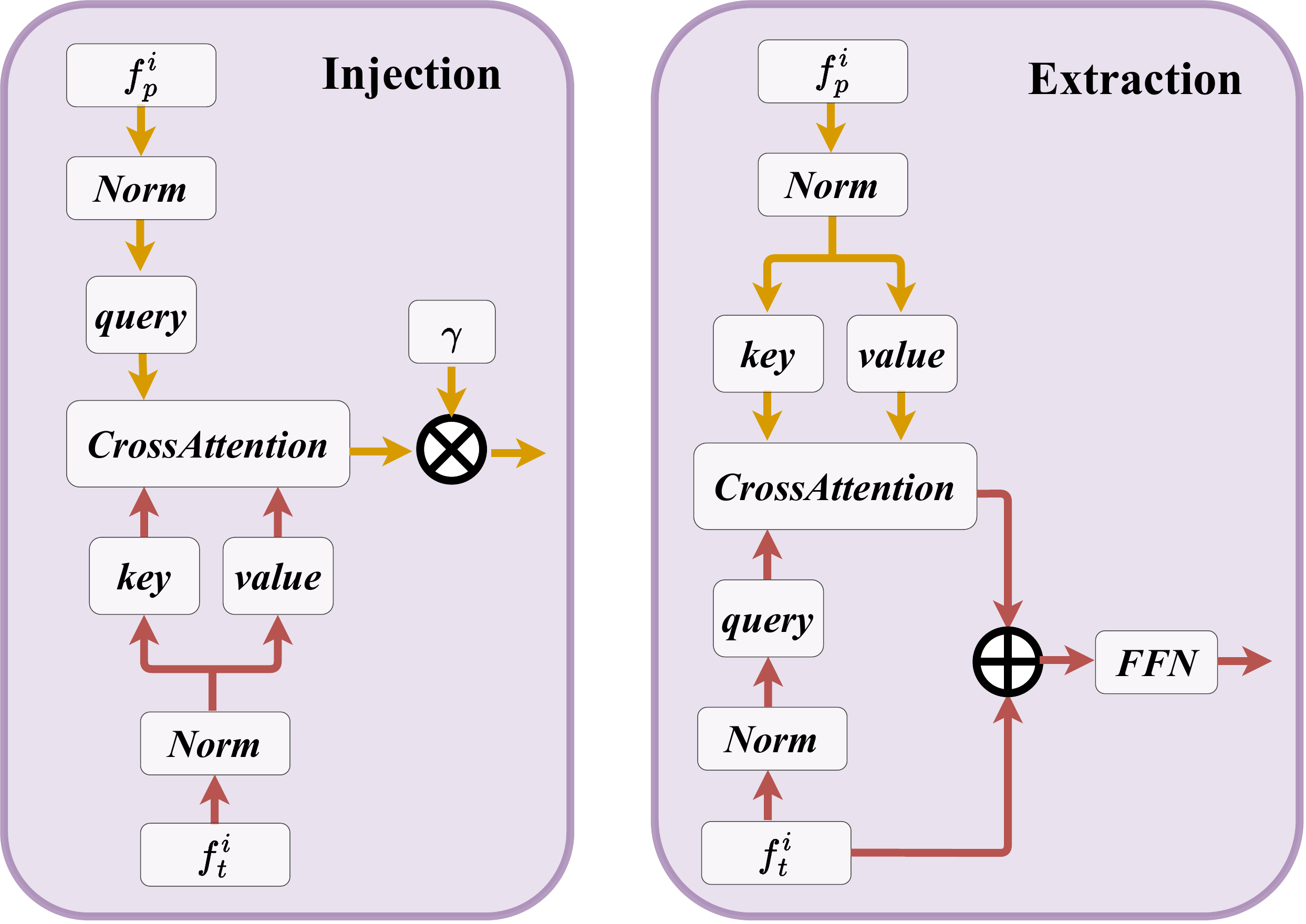}
    \caption{\textbf{Architecture of the Injection and Extraction module.} The left figure shows the details of the injection operation. The right figure displays the structure of the extraction operation.}
    \label{fig:injection_extraction}
\end{figure}

To enhance the interaction between radar features from the two different backbones, we introduce the Injection and Extraction module, which is based on cross-attention, as illustrated in Figure \ref{fig:injection_extraction}. This module is applied at each block of the two backbones. 

Concretely, assuming the features from the $i~th$ block of the point-based and transformer-based backbone are $f_p^i$ and $f_t^i$, respectively.
In injection operation, we take $f_p^i$ as the query and $f_t^i$ as the key and value.
We use multi-head cross-attention to inject transformer feature $f_t^i$ into the point feature $f_p^i$, which can be formulated as follows:
\begin{equation}
    f_p^i = f_p^i + \gamma\times \text{CrossAttention}(LN(f_p^i), LN(f_t^i)), \label{eq:injection}
\end{equation}
where \textit{LN} is the LayerNorm and $\gamma$ is a learnable scaling parameter.

Similarly, the extraction operation extracts the point feature $f_p^i$ with cross-attention for the transformer-based backbone.
The extraction operation is defined as follows:
\begin{equation}
    f_t^i = \text{FFN}(f_t^i + \text{CrossAttention}(LN(f_t^i), LN(f_p^i))), \label{eq:extraction}
\end{equation}
where FFN is the FeedForward Network.
The updated features, $f_p^i$ and $f_t^i$, are sent to the next block of their corresponding backbone.

\begin{figure}
    \centering
    \includegraphics[width=0.95\linewidth]{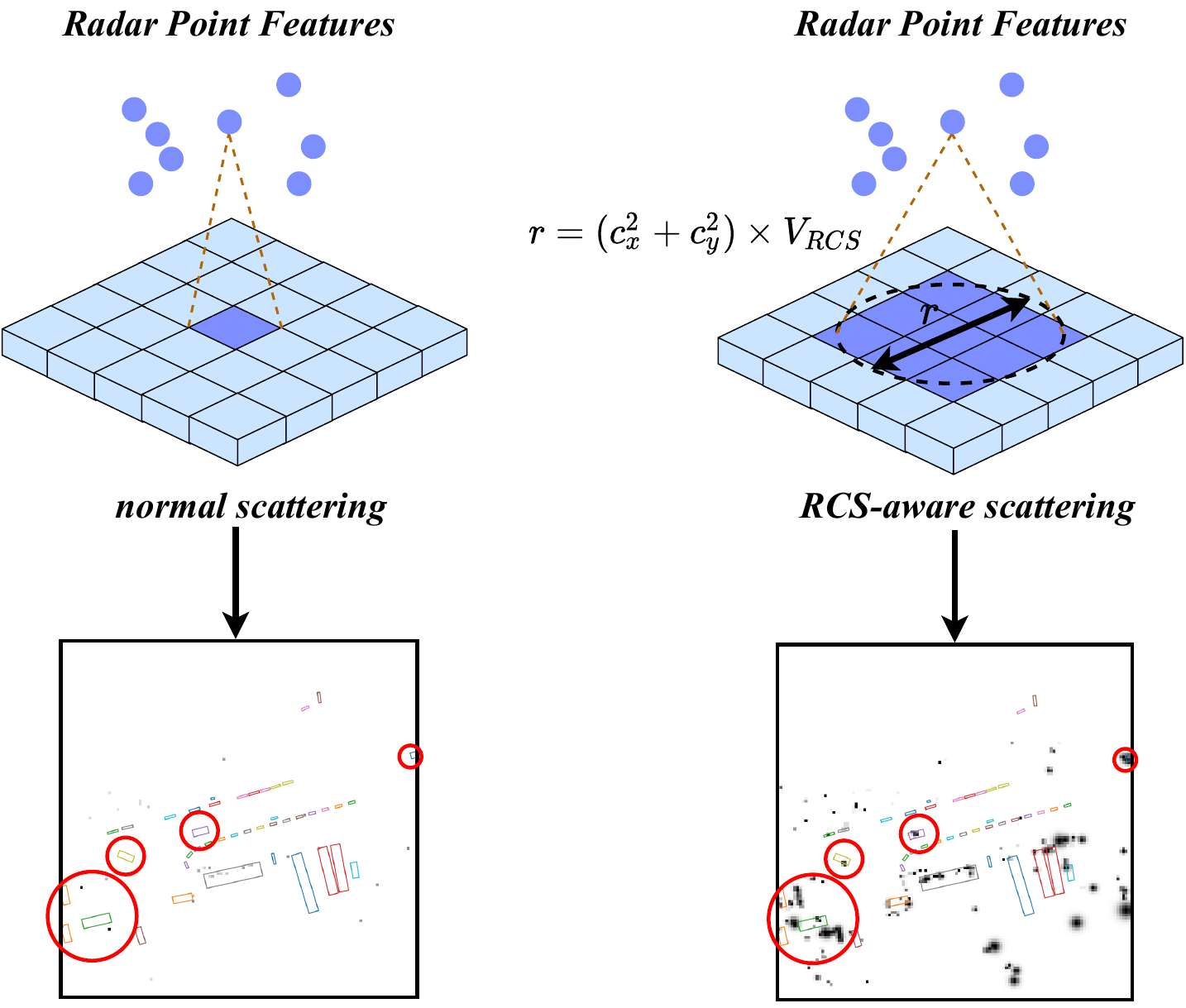}
    \caption{\textbf{Illustration of RCS-aware scattering.} RCS-aware scattering uses RCS as the object size prior to scattering the feature of one radar point to many BEV pixels. 
    }
    \label{fig:rcs}
\end{figure}

\subsubsection{RCS-aware BEV encoder}
Current radar BEV encoders typically scatter point features into BEV space based on the 3D coordinates of the points. 
However, this often results in a sparse BEV feature map, where most pixels contain zero values. This sparsity makes it difficult for some pixels to effectively aggregate features, potentially impairing detection performance. 
One solution is to increase the number of BEV encoder layers, but this can cause small object features to be smoothed out by background features.
To address this issue, we propose an RCS-aware BEV encoder. 
The Radar Cross Section (RCS) measures an object's detectability by radar. For instance, larger objects generally produce stronger radar wave reflections, resulting in a larger RCS measurement. Thus, RCS can provide a rough estimate of an object's size. 
The key design of the RCS-aware BEV encoder is the RCS-aware scattering operation, which leverages the RCS as a prior estimate of the object's size, 
With this prior, the proposed scattering operation allows the feature of a single radar point to be scattered across multiple pixels in the BEV space, rather than being confined to a single pixel, as illustrated in Figure \ref{fig:rcs}.

In particular, without loss of generality, given a specific radar point and its RCS value $V_{RCS}$, 3D coordinate $c=(c_x,c_y)$, BEV pixel coordinate $p=(p_x,p_y)$, and feature $f$, we scatter $f$ to pixel $p$ and nearby pixels, whose pixel distance with $p$ is smaller than $(c_x^2+c_y^2)\times V_{RCS}$.
If a pixel in the BEV feature receives $f$ from multiple radar features, we perform summation pooling to aggregate these features. This operation ensures that all relevant radar information is combined effectively, resulting in a comprehensive radar BEV feature $f_{RCS}$.
Besides, we introduce a Gaussian-like BEV weight map for each point according to the RCS value as follows:
\begin{equation}
    G_{x,y}=\exp(-\frac{(c_x-x)^2+(c_y-y)^2}{\frac{1}{3}(c_x^2+c_y^2)\times V_{RCS}}),
\end{equation}
where $x,y$ are the pixel coordinates.
The final Gaussian-like BEV weight map $G_{RCS}$ is obtained by maximization over all Gaussian-like BEV weight maps.
Subsequently, we concatenate $f_{RCS}$ with $G_{RCS}$ and send them to an MLP to get the final RCS-aware BEV feature as follows:
\begin{equation}
    f_{RCS}'=\text{MLP}(\text{Concat}(f_{RCS},G_{RCS})).
\end{equation}
After that, $f_{RCS}'$ is concatenated with the original BEV feature and sent to the BEV encoder, \textit{e.g.,} SECOND~\cite{yan2018second}.

\label{Sec: BEV cross attention}
\begin{figure}
    \centering
    \includegraphics[width=0.95\linewidth]{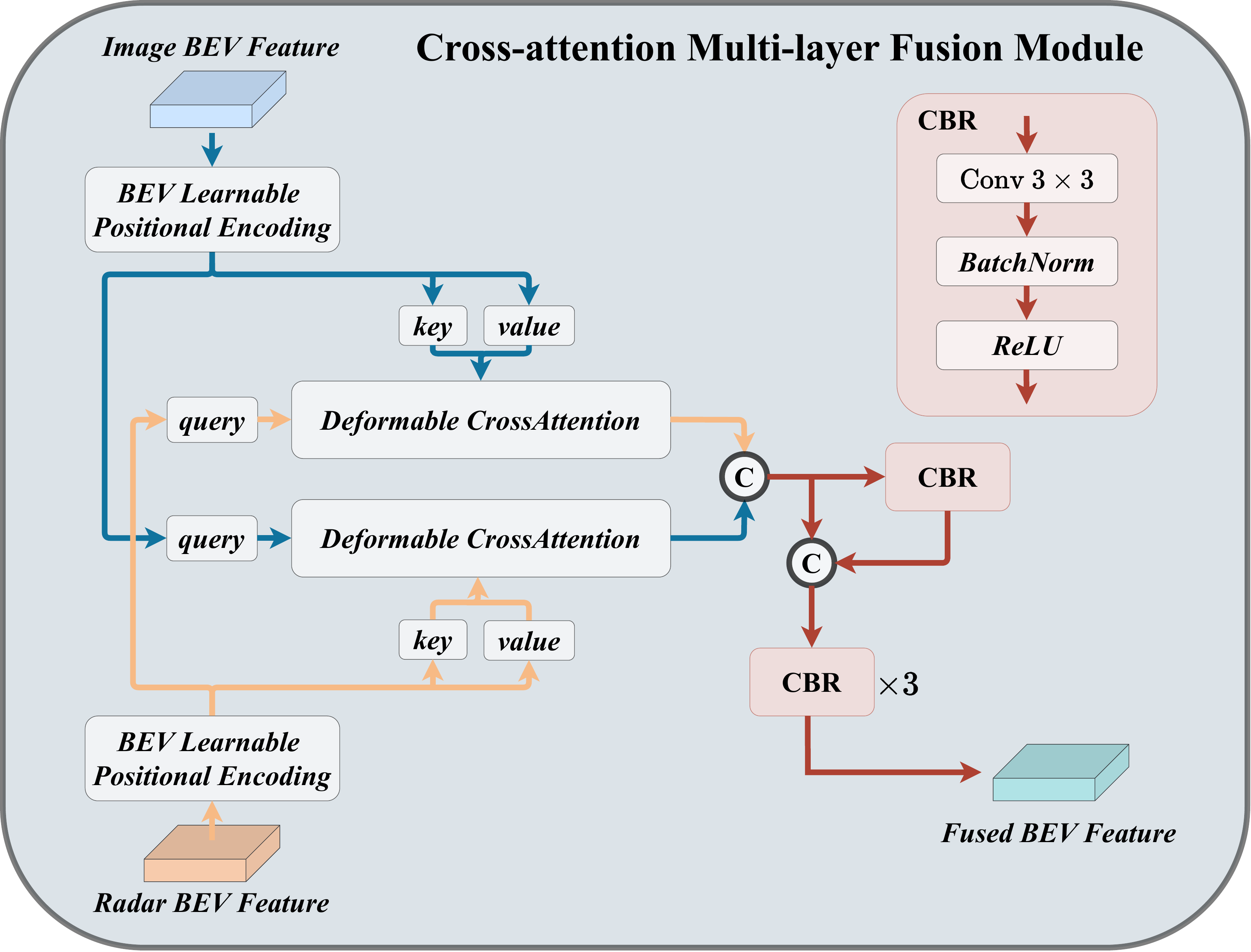}
    \caption{\textbf{Cross-attention multi-layer fusion module.} 
    The BEV features from radar and cameras are dynamically aligned using deformable cross-attention. Subsequently, the multi-modal BEV features are aggregated through a channel and spatial fusion module, which consists of several Convolution-BatchNorm-ReLU (CBR) blocks.
    }
    \label{fig:CAMF}
\end{figure}

\subsection{Cross-Attention Multi-layer Fusion Module}
\subsubsection{Multi-modal Feature Alignment with Cross-Attention}
Radar point clouds often suffer from azimuth errors, causing radar sensors to detect points outside the boundaries of objects. Consequently, radar features generated by RadarBEVNet may be assigned to adjacent BEV grids, resulting in misalignment with BEV features from cameras. 
To address this issue, we use a cross-attention mechanism to dynamically align the multi-modal features. 
Since unaligned radar points typically deviate from their true position by a small distance, we propose employing deformable cross-attention~\cite{zhu2020deformable} to accurately capture and correct these deviations. Besides, the deformable cross-attention can reduce the computational complexity of the vanilla cross-attention from $O(H^{2}W^{2}C)$ to $O(HWC^2K)$, where $H$ and $W$ represent the height and width of the BEV feature, $C$ denotes the BEV feature channels, and $K$ is the number of the reference points in deformable cross-attention.

\begin{figure}[t]
    \centering
    \includegraphics[width=0.95\linewidth]{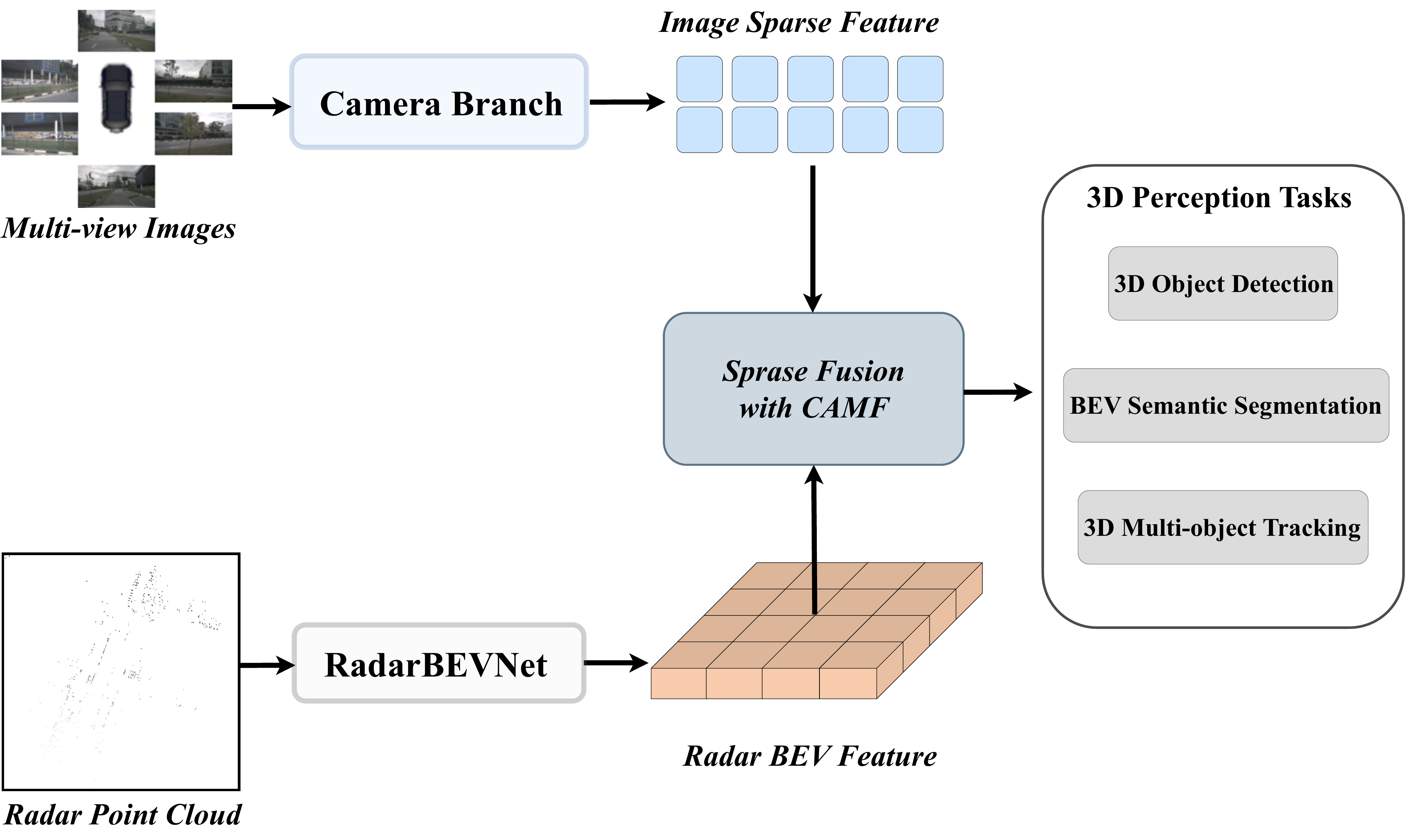}
    \caption{\textbf{Overall pipeline of RCBEVDet++.} 
    The image sparse feature and dense radar BEV feature are fused with the sparse fusion module.
    Then, the fused features are utilized for various 3D perception tasks, including 3D object detection, BEV semantic segmentation, and 3D multi-object tracking.
    }
    \label{fig:rcbevdetpp}
\end{figure}

Specifically, as shown in Figure \ref{fig:CAMF}, given camera and radar BEV features denoted by $ F_{c} \in \mathbb{R}^{C_{c} \times H \times W}$, $F_{r} \in \mathbb{R}^{C_{r} \times H \times W}$, respectively, we first add learnable position embeddings to $F_{c}$ and $F_{r}$.
Then, $F_{r}$ is transformed to queries $z_{q_r}$ and reference points $p_{q_r}$, and $F_{c}$ is viewed as keys and values.
Next, we calculate the multi-head deformable cross-attention~\cite{zhu2020deformable} by:
\begin{equation}
    \begin{aligned}
    &\operatorname{DeformAttn}\left(z_{q_r}, p_{q_r}, F_{c}\right)= \\
    &\sum_{h=1}^{H} \mathrm{~W}_{h}\left[\sum_{k=1}^{K} A_{h q k} \cdot \mathrm{W}_{h}^{\prime} F_c\left(p_{q_r}+\Delta p_{h q k}\right)\right],
\end{aligned}
\end{equation}
where $h$ indexes the attention head, $k$ indexes the sampled keys, $K$ indicates the total sampled key number, $\Delta p_{h q k}$ denotes the sampling offset, $A_{hqk}$ represents attention weight calculated by $z_{q_r}$ and $F_c$, $\mathrm{W}_{h}$ means output weight value to fuse multi-head attention, and $\mathrm{W}_{h}^{\prime}$ is the value projection matrix at $h^{\text{th}}$ head.

Similarly, we exchange $F_{r}$ and $F_{c}$ and conduct another deformable cross-attention to update $F_{r}$. 
Finally, the deformable cross-attention module in CAMF can be formulated as follows:
\begin{equation}
\left\{\begin{aligned}
F_{c} & \leftarrow \operatorname{DeformAttn}\left(z_{q_{r}}, p_{q_{r}}, F_{c}\right), \\
F_{r} & \leftarrow \operatorname{DeformAttn}\left(z_{q_{c}}, p_{q_{c}}, F_{r}\right). \\
\end{aligned}\right.
\end{equation}

\subsubsection{Channel and Spatial Fusion}
After aligning the radar and camera BEV feature by cross-attention, we propose channel and spatial fusion layers to aggregate multi-modal BEV features, as illustrated in Figure \ref{fig:CAMF}.
Specifically, we first concatenate two BEV features as ${F}_{multi}=\left[{F}_{{c}}, {F}_{{r}}\right]$.
Then, ${F}_{multi}$ is sent to a CBR block with a residual connection to obtain the fused feature. 
The CBR block is successively composed of a Conv $3\times3$, a Batch Normalization, and a ReLU activate function.
After that, three CBR blocks are applied to further fuse the multi-modal features.

\section{RCBEVDet++: Radar-camera sparse fusion for 3D perception}
\label{sec:rcbevdet++}
As illustrated in Figure~\ref{fig:rcbevdetpp}, to unleash the full potential of RCBEVDet, we extend the CAMF module to accommodate sparse fusion with query-based multi-view camera perception models, which achieve higher accuracy than BEV-based methods. Besides, we apply RCBEVDet to more perception tasks, including 3D object detection, BEV semantic segmentation, and 3D multi-object tracking. 
To distinguish this updated version of RCBEVDet from the original one, we specially named it RCBEVDet++.

\begin{figure}
    \centering
    \includegraphics[width=0.95\linewidth]{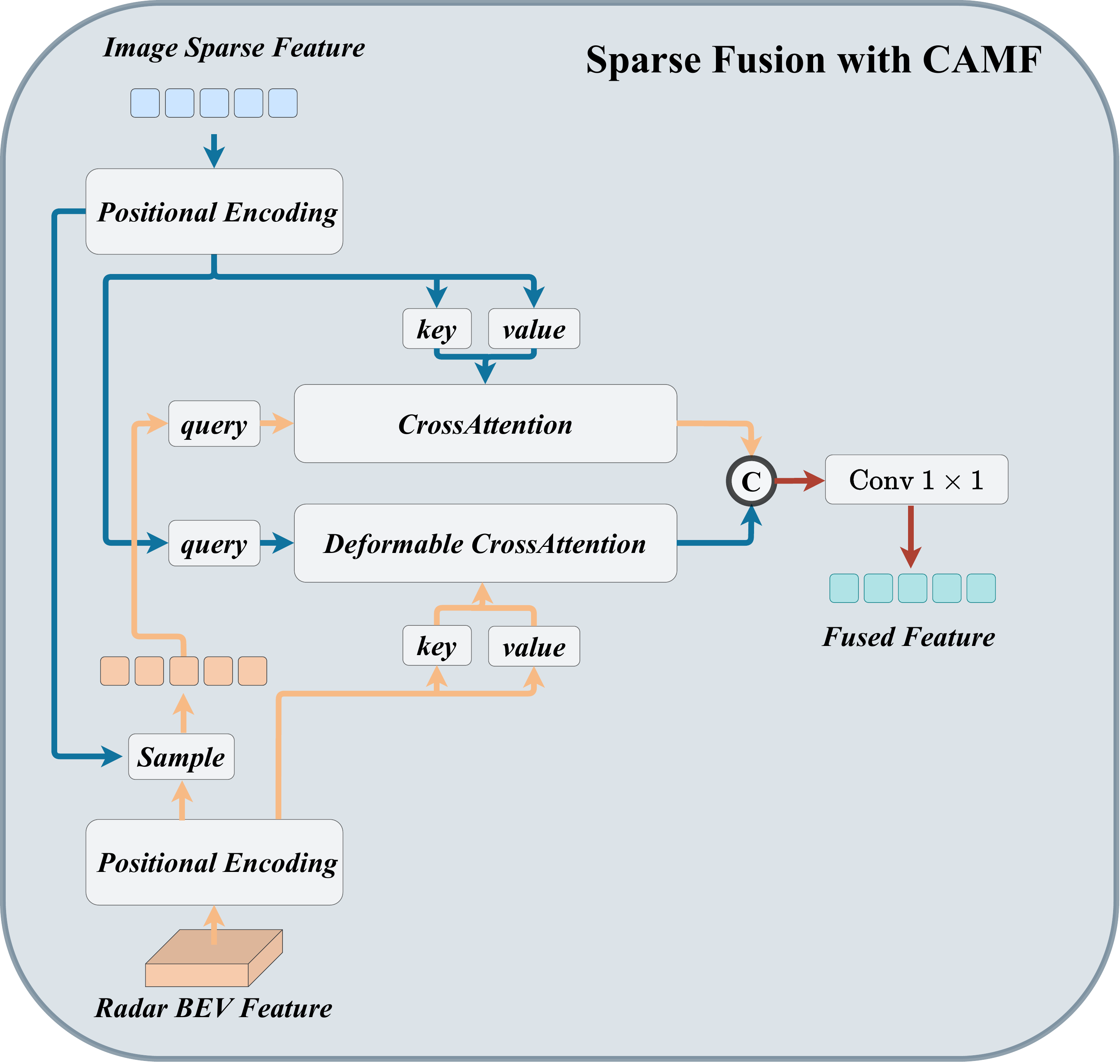}
    \caption{\textbf{Sparse fusion with CAMF.} The dense radar BEV features and image sparse features are dynamically aligned with the cross-attention.
    Then, the multi-modal sparse features are fused with a simple Conv 1$\times$1.
    }
    \label{fig:CAMF_sparse}
\end{figure}

\subsection{Sparse Fusion with CAMF}
As shown in Figure~\ref{fig:CAMF_sparse}, we adopt sparse fusion with CAMF to fuse dense radar BEV features and image sparse features.
Specifically, we first replace the original image BEV features with
image sparse features. 
Then, we perform a project-and-sample process to associate each image sparse feature with a radar feature using 3D absolute position. 
More specifically, we project the 3D absolute position into BEV and sample corresponding radar features with bilinear interpolation to obtain sparse radar features.
Next, we utilize the positional encoding network composed of MLP to transform the 3D absolute position into 3D positional embedding and add them to the multi-modal queries.
After that, to align the multi-modal mismatch, we adopt deformable cross-attention for sparse image features and dense radar BEV features, and simple cross-attention for sparse radar features and sparse image features as follows:
\begin{equation}
\left\{\begin{aligned}
F_{c}^{sparse} & \leftarrow \operatorname{DeformAttn}\left(z_{q_{r}}, p_{q_{r}}, F_{c}^{sparse}\right), \\
F_{r}^{sparse} & \leftarrow \operatorname{CrossAttn}\left(F_{c}^{sparse}, F_{r}\right). \\
\end{aligned}\right.
\end{equation}
where $F^{sparse}$ denotes the sparse feature for radar or image.
Finally, we adopt a simple linear layer to fuse the sparse multi-modal features.

\subsection{Downstream 3D Perception Tasks}
Our RCBEVDet++ can generate high-quality multi-modal features, which can be leveraged for various 3D perception tasks, including 3D object detection, 3D multi-object tracking, and BEV semantic segmentation. To predict the 3D bounding boxes for 3D object detection, we adopt a query-based transformer decoder~\cite{Liu_2023_ICCVSparseBEV} and apply our sparse fusion with the CAMF module in every transformer decoder layer.

After that, we employ the tracking-by-detection framework for the 3D multi-object tracking task. Specifically, we perform velocity-based greedy distance matching; that is, we calculate the center distance of each object in multiple frames with the predicted velocity compensation and assign the same ID for objects with the minimum center distance in a greedy way.

For BEV semantic segmentation, we transform multi-modal features into dense BEV features since this requires a dense BEV map with categories.
We follow the decoder architecture in CVT~\cite{Zhou_2022_CVPRcvt} to effectively decode dense BEV features to different maps with semantic representation.
Additionally, we employ multiple heads to perform different types of BEV semantic segmentation tasks.
Each head deals with one task, \textit{e.g.,} vehicle segmentation.
Finally, focal loss~\cite{lin2017focal} with a sigmoid layer is used as the supervision for training.

\section{Experiments}
In this section, we evaluate RCBEVDet and RCBEVDet++ through extensive experiments. 
In Section~\ref{sec:Implementation Details}, we detail the experimental setup. 
In Section~\ref{sec:sota}, we compare our method with state-of-the-art methods in three tasks, \textit{i.e.}, 3D object detection, BEV semantic segmentation, and 3D multi-object tracking. 
In Section~\ref{sec:ablation}, we conduct an extensive ablation study to investigate individual components of RCBEVDet and RCBEVDet++.
In Section~\ref{sec:bev seg}, we discuss the task trade-off in BEV semantic segmentation of RCBEVDet++.
In Section~\ref{sec:robust}, we show the robustness of RCBEVDet. 
In Section~\ref{sec:generalization}, we demonstrate the model generalization of our method.

\subsection{Implementation Details}
\label{sec:Implementation Details}
\subsubsection{Datasets and Evaluation Metrics}
We conduct experiments on a popular large-scale autonomous diving benchmark, nuScenes~\cite{nuscenes}, which includes 1000 driving scenes collected in Boston and Singapore.
It comprises 850 scenes for training and validation and 150 scenes for testing.
We report the results on the validation and test sets to compare with state-of-the-art methods and evaluate ablation results on the validation set.

For 3D object detection, nuScenes provides a set of evaluation metrics, including mean Average Precision (mAP) and five true positive (TP) metrics: ATE, ASE, AOE, AVE, and AAE, which measure translation, scale, orientation, velocity, and attribute errors, respectively. The overall performance is measured by the nuScenes Detection Score (NDS) that consolidates all error types:
\begin{equation}
    \text{NDS}=\frac{1}{10}[5\times \text{mAP}+\sum_{\text{mTP}\in \text{TP}}(1-\min(1,\text{mTP}))].
\end{equation}

For BEV semantic segmentation, we use mean Intersection over Union (mIoU) across all segmentation categories as the metric, following the settings of LSS~\cite{philion2020liftLSS}.

For 3D multi-object tracking, we follow the official nuScenes metrics, which use Average Multi-Object Tracking Accuracy (AMOTA) and Average Multi-Object Tracking Precision (AMOTP) over various recall thresholds. Concretely, AMOTA is defined as follows:
\begin{equation}
AMOTA=\frac{1}{n-1}\sum_{r\in\{\frac{1}{n-1},\frac{2}{n-1} \dots 1\}}MOTAR_{r},
\end{equation}
\begin{equation}
\begin{aligned}
&MOTAR_{r}= \\
&\max\left(0, 1 - \frac{IDS_r+FP_r+FN_r-(1-r)P}{rP}\right),
\end{aligned}
\end{equation}
where $P$ refers to the number of true positives of the current class, $r$ is the scalar factor, $IDS$ represents the number of identity switches, $FP$ and $FN$ denote the number of false positives and false negatives, respectively, and $n$ is set to 40.
For AMOTP, it can be formulated as follows:
\begin{equation}
AMOTP=\frac{1}{n-1}\sum_{r\in\{\frac{1}{n-1},\frac{2}{n-1},...,1\}}\frac{\sum_{i,t}d_{i,t}}{\sum_{t}TP_{t}},
\end{equation}
where $d_{i,t}$ represents the position error of tracked object $i$ at time $t$ and $TP_t$ indicates the number of matches at time $t$.

\subsubsection{Architecture and Training Details}
We adopt BEVDepth~\cite{li2023bevdepth} with BEVPoolv2~\cite{huang2022bevpoolv2} and SparseBEV~\cite{Liu_2023_ICCVSparseBEV} as the camera stream for RCBEVDet and RCBEVDet++, respectively. 
For BEVDepth, we follow BEVDet4D~\cite{huang2022bevdet4d} to accumulate the intermediate BEV feature from multiple frames and add an extra BEV encoder to aggregate these multi-frame BEV features.
For radar, we accumulate multi-sweep radar points and use RCS and Doppler speed as input features in the same manner as GRIFNet~\cite{kim2020grif} and CRN~\cite{Kim_2023_ICCVCRN}. 
We set the number of stages $S$ in the dual-stream radar backbone to 3.

For the 3D object detection head, we use the center head from CenterPoint~\cite{yin2021centerpoint} for RCBEVDet and the sparse head from SpraseBEV~\cite{Liu_2023_ICCVSparseBEV} for RCBEVDet++.
For the BEV semantic segmentation head, we adopt a separate segmentation head for each task.
For 3D multi-object tracking, we follow CenterPoint to utilize estimated velocity to track object centers across multiple frames in a greedy manner.

Our models are trained in a two-stage manner.
In the first stage, we train the camera-based model following the standard implementations~\cite{li2023bevdepth, Liu_2023_ICCVSparseBEV}.
In the second stage, we train the radar-camera fusion model.
The weights of the camera stream are inherited from the first stage, and the parameters of the camera stream are frozen during the second stage.
All models are trained for 12 epochs with AdamW~\cite{kingma2015adam} optimizer.
To prevent overfitting, we apply various data augmentations, including image rotation, cropping, resizing, and flipping, as well as radar horizontal flipping, horizontal rotation, and coordinate scaling.
%
%


\setlength{\tabcolsep}{0.65em}
\begin{table*}[!t]
\caption{
    \textbf{Comparison of 3D object detection results on nuScenes \texttt{val} set}. 
    `C' and `R' represent camera and radar, respectively.
}
\begin{center}
\label{table:3D det val set}
\resizebox{0.98\textwidth}{!}{
\begin{tabular}{l|c|c|c||cc|ccccc}
    \hline
    Method & Input & Backbone & Image Size & NDS$\uparrow$ & mAP$\uparrow$ & mATE$\downarrow$ & mASE$\downarrow$ & mAOE$\downarrow$ & mAVE$\downarrow$ & mAAE$\downarrow$ \\
    \hline
    CenterFusion \cite{nabati2021centerfusion}
    & C+R & DLA34 & $448\times800$  & 45.3 & 33.2 & 0.649 & \textbf{0.263} & 0.535 & 0.540 & \textbf{0.142} \\    
    CRAFT \cite{kim2023craft}
    & C+R & DLA34 & $448\times800$  & 51.7 & 41.1 & 0.494 & 0.276 & 0.454 & 0.486 & 0.176  \\
    \gr \ours~(Ours) & C+R & DLA34 &  $448\times800$ & \textbf{56.3} & \textbf{45.3} & \textbf{0.492} & 0.269 & \textbf{0.449} & \textbf{0.230} & 0.188 \\
    \hline
    RCBEV4d \cite{RCBEV}
    & C+R & Swin-T & $256\times704$ & 49.7 & 38.1 & 0.526 & 0.272 & 0.445 & 0.465 & 0.185  \\
    \gr \ours~(Ours) & C+R & Swin-T &  $256\times704$ & \textbf{56.2} & \textbf{49.6}  & \textbf{0.496} &\textbf{0.271} & \textbf{0.418} & \textbf{0.239} & \textbf{0.179} \\
    \hline
    CRN~\cite{Kim_2023_ICCVCRN} & C+R& ResNet-18 & $256\times704$ & 54.3& \textbf{44.8}& 0.518& \textbf{0.283}& 0.552& 0.279& 0.180 \\
    \gr \ours~(Ours) & C+R & ResNet-18 &  $256\times704$ &  \textbf{54.8} & 42.9 & \textbf{0.502}& 0.291& \textbf{0.432}& \textbf{0.210}& \textbf{0.178}  \\
    \hline
    BEVDet \cite{huang2021bevdet}
    & C & ResNet-50 & $256\times704$& 39.2 & 31.2 & 0.691 & 0.272 & 0.523 & 0.909 & 0.247 \\
    BEVDepth \cite{li2023bevdepth}
    & C & ResNet-50 & $256\times704$ & 47.5 & 35.1 & 0.639 & \textbf{0.267} & 0.479 & 0.428 & 0.198  \\
    SOLOFusion \cite{DBLP:conf/iclr/solofusion}
    & C & ResNet-50 & $256\times704$ & 53.4 & 42.7 & 0.567 & 0.274 & {0.411} & {0.252} & {0.188} \\
    StreamPETR~\cite{wang2023exploringStreamPETR} & C & ResNet-50 & $256\times704$ & 54.0 & 43.2 & 0.581 & 0.272 & 0.413 & 0.295 & 0.195 \\
    SparseBEV~\cite{Liu_2023_ICCVSparseBEV} & C & ResNet-50 & $256\times704$ & 54.5 & 43.2& 0.606& 0.274 &0.387 &0.251 & 0.186\\
    
    {CRN}~\cite{Kim_2023_ICCVCRN} 
    & C+R & ResNet-50 & $256\times704$&  {56.0} & {49.0} & {0.487} & 0.277 & 0.542 & 0.344 & 0.197 \\
    \gr \ours~(Ours) & C+R & ResNet-50 &  $256\times704$ & {56.8} & 45.3 & \textbf{0.486}  & 0.285 & \textbf{0.404} & \textbf{0.220} & 0.192 \\
    \gr \ours++~(Ours) & C+R & ResNet-50 &  $256\times704$ & \textbf{60.4} & \textbf{51.9} & {0.488}  & 0.268 & {0.408} & {0.221} & \textbf{0.177} \\

    \hline
\end{tabular}}
\end{center}
\end{table*}

\renewcommand{\thefootnote}{1}
\setlength{\tabcolsep}{0.65em}
\begin{table*}[!t]
\begin{center}
\caption{
    \textbf{Comparison of 3D object detection results on nuScenes \texttt{test} set}.
    `L', `C', and `R' represent LiDAR, camera, and radar, respectively.
    $^{\dagger}$ uses future frames.
    We do not use test-time data augmentation or ensemble for submission.
}
\label{table:3D det test set}
\resizebox{0.98\textwidth}{!}{
\begin{tabular}{l|c|c||cc|ccccc}
    \hline
    Method & Input  & Backbone & NDS$\uparrow$ & mAP$\uparrow$ & mATE$\downarrow$ & mASE$\downarrow$ & mAOE$\downarrow$ & mAVE$\downarrow$ & mAAE$\downarrow$\\
    \hline
    CenterPoint~\cite{yin2021centerpoint} & L & VoxelNet & 67.3 & 60.3 & 0.262 &0.239&0.361&0.288&0.136 \\
    TransFusion-L~\cite{bai2022transfusion} & L & VoxelNet &  70.2 & 65.5 & 0.256 & 0.240 &0.351 &0.278 &0.129 \\
    \hline
    {KPConvPillars} \cite{KPConvPillars}   & R & Pillars    & 13.9 &  4.9 & 0.823 & 0.428 &0.607&2.081&1.000 \\
    \hline
    CenterFusion \cite{nabati2021centerfusion}  &C+R& DLA34    & 44.9 & 32.6 & 0.631 & 0.261 & 0.516 &0.614&0.115 \\
    
    RCBEV \cite{RCBEV}               &C+R& Swin-T   & 48.6 & 40.6 & 0.484 & 0.257 & 0.587& 0.702 &0.140 \\
    
    
    MVFusion \cite{DBLP:conf/icra/mvfusion}              &C+R& V2-99    & 51.7 & 45.3 & 0.569 &  0.246 & 0.379& 0.781& 0.128 \\
    
    CRAFT \cite{kim2023craft}                   &C+R& DLA34    & 52.3 & 41.1 & 0.467 & 0.268 & 0.456 &0.519&0.114 \\
    
    BEVFormer \cite{li2022bevformer}            & C & V2-99    & 56.9 & 48.1 & 0.582 & 0.256 &0.375& 0.378& 0.126 \\
    PETRv2~\cite{liu2023petrv2} & C & V2-99 & 58.2 & 49.0 & 0.561& 0.243& 0.361& 0.343& 0.120 \\
    BEVDepth \cite{li2023bevdepth}              & C & V2-99 & 60.5 & 51.5 & 0.446&  0.242& 0.377& 0.324& 0.135 \\
    BEVDepth \cite{li2023bevdepth}              & C & {ConvNeXt-B} & 60.9 & 52.0 & 0.445&  0.243& 0.352& 0.347& 0.127 \\
    
    BEVStereo \cite{li2019stereo}            & C & V2-99    & 61.0 & 52.5 & 0.431 & 0.246 &0.358& 0.357& 0.138 \\
    
    SOLOFusion \cite{DBLP:conf/iclr/solofusion}              & C & {ConvNeXt-B} & 61.9 & 54.0 & 0.453 & 0.257& 0.376 &0.276& 0.148 \\
    
    {CRN}~\cite{Kim_2023_ICCVCRN}                            &C+R& {ConvNeXt-B} & {62.4} & {57.5} & 0.416 & 0.264 & 0.456 & 0.365 & 0.130 \\
    {StreamPETR}~\cite{wang2023exploringStreamPETR}                            &C& {V2-99} &63.6 & 55.0  &0.493& 0.241 &{0.343}& {0.243}& 0.123 \\
    SparseBEV~\cite{Liu_2023_ICCVSparseBEV} & C & V2-99& 63.6 &55.6 &0.485& 0.244& {0.332}& 0.246& 0.117\\
    SparseBEV$^{\dagger}$~\cite{Liu_2023_ICCVSparseBEV} & C & V2-99& 67.5 & 60.3 & 0.425 & 0.239 & 0.311 & 0.172 & 0.116\\
    \gr \ours~(Ours)                          &C+R& {V2-99} & {63.9} & 55.0 & \textbf{0.390} &\textbf{0.234} & 0.362 & 0.259 & \textbf{0.113} \\
    
    \gr \ours++~(Ours)                          &C+R& {V2-99} & \textbf{68.7} & \textbf{62.6} & {0.437} &{0.252} & \textbf{0.181} & \textbf{0.203} & {0.191} \\

    \hline
    {StreamPETR}~\cite{wang2023exploringStreamPETR}                            &C& {ViT-L} & 67.6 & 62.0 & 0.470 & 0.241 & 0.258 & 0.236 & 0.134 \\
    Far3D~\cite{jiang2024far3d}                            &C& {ViT-L} & 68.7& 63.5 & 0.432 & 0.237 & 0.278 & 0.227 & \textbf{0.130} \\
    {SparseBEV}$^{\dagger}$~\cite{wang2023exploringStreamPETR}                            &C& {ViT-L} & 70.2 & - & - & - & - & - & - \\
    \gr \ours++$^{\dagger}$~(Ours)                          &C+R& {ViT-L} & \textbf{72.7} & \textbf{67.3} & \textbf{0.341} &\textbf{0.234} & \textbf{0.241} & \textbf{0.147} & \textbf{0.130} \\
    \hline

\end{tabular}}
\end{center}
\end{table*}

\subsection{Comparison with State-of-the-Art}
\label{sec:sota}
We compare our methods with cutting-edge camera-based and radar-camera multi-modal methods in three tasks: 3D object detection, BEV semantic segmentation, and 3D multi-object tracking.

\subsubsection{3D Object Detection}
We provide 3D object detection results on \texttt{val} and \texttt{test} set in Tables~\ref{table:3D det val set} and~\ref{table:3D det test set}, respectively. 

As shown in Table~\ref{table:3D det val set}, RCBEVDet outperforms previous radar-camera multi-modal 3D object detection methods across various backbones. Besides, based on SparseBEV, RCBEVDet++ surpasses CRN by 4.4 NDS, showing the effectiveness of our fusion method.
Furthermore, RCBEVDet and RCBEVDet++ reduce the velocity error by 14.6\% compared with the previous best method, demonstrating the efficiency of our method in utilizing radar information.

On \texttt{test} sets, with the V2-99 backbone, RCBEVDet++ improves the SparseBEV baseline by 5.1 NDS and 7.0 mAP and surpasses its offline version with future frames.
Notably, RCBEVDet++ with the smaller V2-99 backbone achieves competitive performance compared to StreamPETR and Far3D with the larger backbone ViT-L.
Moreover, RCBEVDet++ with the larger ViT-L backbone achieves 72.7 NDS and 67.3 mAP without test-time data augmentation, setting new state-of-the-art radar-camera 3D object detection results on nuScenes.

\subsubsection{BEV Semantic Segmentation}
We compare our method with state-of-the-art BEV semantic segmentation methods on \texttt{val} set in Tables~\ref{table:seg}. 
With the ResNet-101 backbone, RCBEVDet++ shows a favorable performance increase over CRN for the “Drivable Area” class (+0.6 IoU) and over BEVGuide for the “Lane” class (+6.3 IoU), respectively.
In the combined evaluation for all tasks, RCBEVDet++ achieves state-of-the-art performance with 62.8 mIoU, outperforming previous best results by 1.8 mIoU.
These results demonstrate the effectiveness of our method in dealing with the BEV semantic segmentation task.

\renewcommand{\thefootnote}{1}
\setlength{\tabcolsep}{0.65em}
\begin{table*}[!t]
\begin{center}
\caption{
    \textbf{Comparison of BEV semantic segmentation on nuScenes \texttt{val} set}.
    `C' and `R' represent camera and radar, respectively.
    `Driv. Area' denotes drivable area.
    `mIoU' is the average IoU over Vehicle, Drivable Area, and Lane.
    $^{\dagger}$ is a Simple-BEV~\cite{Harley2023Simple-BEV} customized by BEVCar~\cite{schramm2024bevcar}.
}
\label{table:seg}
\begin{tabular}{l|cc||c|ccc}
    \hline
    Methods & Input & Backbone& mIoU$\uparrow$  & Vehicle$\uparrow$ & Driv. Area$\uparrow$ & Lane$\uparrow$ \\
    \hline
    CVT~\cite{Zhou_2022_CVPRcvt} & C & EfficientNet & 46.6& 36.0 & 74.3 & 29.4  \\
    BEVFormer-S~\cite{li2022bevformer} & C & ResNet-101& 48.4 & 43.2 & 80.7 & 21.3 \\
    \hline
    CRN~\cite{Kim_2023_ICCVCRN} & C+R & ResNet-50& - & 58.8 & 82.1 & -  \\
    Simple-BEV++$^{\dagger}$~\cite{schramm2024bevcar} & C+R & ResNet-101& 55.4 & 52.7 & 77.7 & 35.8 \\
    BEVGuide~\cite{Man_2023_CVPRBEVguided} & C+R & EfficientNet& 60.0 & \textbf{59.2} & 76.7 & 44.2  \\
    BEVCar~\cite{schramm2024bevcar} & C+R & ResNet-101& 61.0 & 57.3 & 81.8 & 43.8  \\
    \gr RCBEVDet++ (Ours) & C+R & ResNet-101& \textbf{62.8} & 55.3 & \textbf{82.7} & \textbf{50.5} \\ 
    \hline

\end{tabular}
\end{center}
\end{table*}

\subsubsection{3D Multi-Object Tracking}
In Table~\ref{table:track}, we summarize the 3D multi-object tracking results on nuScenes \texttt{test} set. 
Due to the high accuracy of our method in estimating objects' locations and velocities, RCBEVDet++ achieves the best AMOTA and AMOTP simultaneously compared with state-of-the-art methods.

\setlength{\tabcolsep}{0.9em}
\begin{table}[!t]
\begin{center}
\caption{
    \textbf{Comparison of 3D multi-object tracking on nuScenes \texttt{test} set}.
    `L', `C', and `R' represent LiDAR, camera, and radar, respectively.
}
\label{table:track}
\begin{tabular}{l|c||cc}
    \hline
    Methods & Input & AMOTA$\uparrow$ & AMOTP$\downarrow$ \\
    \hline
    CenterPoint~\cite{yin2021centerpoint} & L & 63.8 & 0.555 \\
    \hline


UVTR~\cite{li2022unifying} & C & 51.9 & 1.125\\
Sparse4D~\cite{Sparse4D} & C & 51.9 & 1.078 \\

ByteTrackV2~\cite{stadler2022bytev2} & C & 56.4 & 1.005  \\
StreamPETR~\cite{wang2023exploringStreamPETR} & C & 56.6 & 0.975 \\
CRN~\cite{Kim_2023_ICCVCRN} & C+R & 56.9 & 0.809 \\
\gr RCBEVDet++ (Ours) & C+R & \textbf{59.6} & \textbf{0.713} \\
    \hline

\end{tabular}
\end{center}
\end{table}

\subsection{Ablation Studies}
\label{sec:ablation}
We ablate various design choices for our proposed method. 
For simplicity, we conduct ablation on the 3D detection task.
All results in the ablation are on the nuScenes
\texttt{val} set with ResNet-50 backbone, $256\times 704$ image input size, and $128\times 128$ BEV size if not specified.

\subsubsection{Main Components}
In this study, we conduct experiments to evaluate the effectiveness of the main components in Section~\ref{sec:rcbevdet}, including RadarBEVNet and CAMF. 
%
Specifically, as shown in Table~\ref{table:main component}, we successively add components to the baseline BEVDepth to compose RCBEVDet.
Firstly, based on the camera-only model, we build a radar-camera 3D object detection baseline with BEVFusion~\cite{DBLP:journals/corr/bevfusionmit} by adopting PointPillar as the radar backbone following CRN~\cite{Kim_2023_ICCVCRN}.
The baseline radar-camera detector achieves 53.6 NDS and 42.3 mAP, improving the camera-only detector by 1.7 NDS and 1.8 mAP.
%
Next, substituting PointPillar with the proposed RadarBEVNet yields 2.1 NDS and 3.0 mAP improvement, demonstrating RadarBEVNet's strong radar feature representation capability. Furthermore, integrating CAMF boosts 3D detection performance from 55.7 NDS to 56.4 NDS. 
Additionally, we follow Hop~\cite{Zong_2023_hop} to incorporate extra multi-frame losses, termed Temporal Supervision, resulting in a 0.4 NDS improvement alongside a 0.3 mAP reduction.

Overall, we observe that each component consistently improves 3D object detection performance. Meanwhile, the results demonstrate that multi-module fusion can significantly boost the detection performance.

\setlength{\tabcolsep}{0.6em}
\begin{table}[!t]
\caption{
\textbf{Ablation of the main components of \ours}. We successively add components to BEVDepth~\cite{li2023bevdepth} to compose \ours. Each component improves the 3D detection performance consistently.
}
\begin{center}
\label{table:main component}
\resizebox{0.98\columnwidth}{!}{
\begin{tabular}{l|c||cc}
\hline
Model Configuration & Input & NDS$\uparrow$ & mAP$\uparrow$\\
\hline
BEVDepth~\cite{li2023bevdepth} & C & 47.5 & 35.1 \\ 
+ Temporal~\cite{huang2022bevdet4d}  & C & 51.9 \textcolor{red}{$\uparrow$4.4} & 40.5 \textcolor{red}{$\uparrow$5.4} \\
\hline
+ PointPillar+BEVFusion  & C+R & 53.6 \textcolor{red}{$\uparrow$1.7} & 42.3 \textcolor{red}{$\uparrow$1.8} \\
+ RadarBEVNet  & C+R & 55.7 \textcolor{red}{$\uparrow$2.1} & 45.3 \textcolor{red}{$\uparrow$3.0} \\
+ CAMF  & C+R & 56.4 \textcolor{red}{$\uparrow$0.7} & \textbf{45.6} \textcolor{red}{$\uparrow$0.3} \\
+ Temporal Supervision  & C+R & \textbf{56.8} \textcolor{red}{$\uparrow$0.4} & 45.3 \textcolor{green}{$\downarrow$0.3} \\
\hline
\end{tabular}
}
\end{center}
\end{table}

\setlength{\tabcolsep}{0.6em}
\begin{table}[!t]
\caption{
\textbf{Ablation of RadarBEVNet}. 
The dual-stream radar backbone obtains marginal performance gains without the Injection and Extraction module.
}
\begin{center}
\label{table:RadarBEVNet}
\begin{tabular}{l||cc}
\hline

Radar Backbone & NDS$\uparrow$ & mAP$\uparrow$\\
\hline
PointPillar & 54.3 & 42.6 \\ 
+ RCS-aware BEV encoder   & 55.7 \textcolor{red}{$\uparrow$1.4} & 44.5 \textcolor{red}{$\uparrow$1.9} \\
\hline
+ Transformer Backbone  & 55.8 \textcolor{red}{$\uparrow$0.1} & 44.8 \textcolor{red}{$\uparrow$0.3} \\
+ Injection and Extraction module  & \textbf{56.4} \textcolor{red}{$\uparrow$0.6} & \textbf{45.6} \textcolor{red}{$\uparrow$0.8} \\
\hline
\end{tabular}
\end{center}
\end{table}

\subsubsection{RadarBEVNet}
Experimental results related to the design of RadarBEVNet, including the Dual-stream radar backbone and RCS-aware BEV encoder, are presented in Table \ref{table:RadarBEVNet}.
Specifically, the baseline model, which uses PointPillar as the radar backbone, achieves 54.3 NDS and 42.6 mAP. 
Integrating the RCS-aware BEV encoder enhances 3D object detection performance by 1.4 NDS and 1.9 mAP, demonstrating the effectiveness of the proposed RCS-aware BEV feature reconstruction.
Additionally, we find that the direct integration of the Transformer-based Backbone leads to marginal performance improvement. 
This is attributed to the individual processing of radar points by the Point-based and Transformer-based backbones, which lack the effective interaction of their distinct radar feature representations. 
To alleviate this issue, we introduce the Injection and Extraction module, resulting in a performance gain of 0.6 NDS and 0.8 mAP.

Furthermore, we compare the proposed RadarBEVNet with PointPillar on different input modalities. As shown in Table~\ref{table:radar}, RadarBEVNet shows superior detection performance compared to PointPillar, with improvements of 1.7 NDS and 2.1 NDS for radar and radar-camera inputs, respectively.

\setlength{\tabcolsep}{0.6em}
\begin{table}[!t]
\caption{
\textbf{Comparison between PointPillar and RadarBEVNet on
different modal inputs}. We evaluate the effectiveness of RadarBEVNet with two modal inputs.
}
\begin{center}
\label{table:radar}
\begin{tabular}{l|c||cc}
\hline

Radar Backbone & Input &NDS$\uparrow$ & mAP$\uparrow$\\
\hline
PointPillar & \multirow{2}{*}{R}& 20.3 &7.3 \\ 
RadarBEVNet  & &\textbf{22.0} \textcolor{red}{$\uparrow$1.7} & \textbf{10.8} \textcolor{red}{$\uparrow$3.5} \\
\hline
PointPillar & \multirow{2}{*}{C+R}& 54.3 & 42.6 \\ 
RadarBEVNet  & &\textbf{56.4} \textcolor{red}{$\uparrow$2.1} & \textbf{45.6} \textcolor{red}{$\uparrow$3.0} \\
\hline
\end{tabular}
\end{center}
\end{table}

\subsubsection{Cross-attention Multi-layer Fusion (CAMF)}
In this study, we conducted ablation experiments on the CAMF module, which includes the deformable cross-attention mechanism for aligning multi-modal features and the channel and spatial fusion module, as shown in Table \ref{table:CAMF}.
Specifically, the baseline model with the fusion module from BEVfusion~\cite{DBLP:journals/corr/bevfusionmit} achieves 55.7 NDS and 45.3 mAP.
When the Deformable Cross Attention is incorporated for multi-modal BEV feature alignment, the 3D detection performance improves from 55.7 NDS and 45.3 mAP to 56.1 NDS and 45.5 mAP. This highlights the effectiveness of the cross-attention mechanism in aligning cross-modal features.
Additionally, we notice that introducing the channel and spatial fusion module for BEV feature fusion obtains a performance increase of 0.3 NDS and 0.1 mAP compared to the one-layer fusion in BEVFusion~\cite{DBLP:journals/corr/bevfusionmit}.
This indicates that channel and spatial multi-layer fusion provides better multi-modal BEV features. 

\setlength{\tabcolsep}{0.6em}
\begin{table}[!t]
\caption{
\textbf{Ablation of CAMF}. 
}
\begin{center}
\label{table:CAMF}
\begin{tabular}{l||cc}
\hline

Fusion Method & NDS$\uparrow$ & mAP$\uparrow$\\
\hline

BEVFusion~\cite{DBLP:journals/corr/bevfusionmit} & 55.7 & 45.3 \\ 
\hline
+ Deformable Cross-Attention  & 56.1 \textcolor{red}{$\uparrow$0.4} & 45.5 \textcolor{red}{$\uparrow$0.2} \\
+ Channel and Spatial Fusion  & \textbf{56.4} \textcolor{red}{$\uparrow$0.3} & \textbf{45.6} \textcolor{red}{$\uparrow$0.1} \\
\hline
\end{tabular}
\end{center}
\end{table}

\subsubsection{Sparse Fusion with CAMF}
Table \ref{table:sparse CAMF} presents the ablation for sparse fusion with CAMF. 
The first row in Table \ref{table:sparse CAMF} denotes the SparseBEV baseline.
%
When only adopting deformable attention to align radar BEV features with image sparse features obtains performance gains of 1.2 NDS and 2.3 mAP.
After adding the radar query sample for multi-modal feature alignment further boosts detection performance by 2.4 NDS and 4.2 mAP.
%
Besides, we observe that replacing learnable positional encoding with non-parametric encoding (\textit{i.e.}, sine positional encoding) improves results by 1.9 NDS and 1.9 mAP.
Lastly, unlike CAMF in RCBEVDet, linear fusion outperforms multi-layer fusion (MLP in Table \ref{table:sparse CAMF}).
The reason is that the BEV features are 2D dense features, which require spatial and channel fusion.
In contrast, sparse query features are 1D features; thus, a linear fusion layer is adequate for practice.

\setlength{\tabcolsep}{0.6em}
\begin{table*}[!t]
\caption{
\textbf{Ablation of sparse fusion with CAMF}. 
}
\begin{center}
\label{table:sparse CAMF}
\begin{tabular}{c|c|c|c||cc}
\hline

DeformAttn-only & Radar Query Sample & Positional Encoding& Fusion Layer & NDS$\uparrow$ & mAP$\uparrow$\\
\hline
$\times$ & $\times$ & - & - & 54.5 & 43.2\\ 
$\checkmark$ & $\times$ & Learnable & MLP & 55.7\textcolor{red}{$\uparrow$1.2} & 45.5\textcolor{red}{$\uparrow$2.3}\\ 
$\checkmark$ & $\checkmark$ & Learnable & MLP & 58.1\textcolor{red}{$\uparrow$2.4} & 49.7\textcolor{red}{$\uparrow$4.2} \\ 
$\checkmark$ & $\checkmark$ & Sine & MLP & 60.0\textcolor{red}{$\uparrow$1.9} & 51.6\textcolor{red}{$\uparrow$1.9} \\ 
$\checkmark$ & $\checkmark$ & Sine & Linear & 60.4\textcolor{red}{$\uparrow$0.4} & 51.9\textcolor{red}{$\uparrow$0.3} \\ 
\hline
\end{tabular}
\end{center}
\end{table*}

\subsection{Task Trade-off in BEV semantic segmentation}
\label{sec:bev seg}
BEV semantic segmentation in nuScenes needs to complete three tasks, including vehicle, drivable area, and lane segmentation.
To achieve an optimal balance among these tasks, we adjust the loss weights of the three tasks and present the results in Table \ref{table:seg trade-off}.
We observe that applying equal loss weights for each task obtains 57.7 mIoU.
By progressively increasing the loss weights of vehicle and lane classes while decreasing the loss weights for the drivable area, segmentation performance first improves from 57.7 mIoU to 59.5 mIoU, reaching its peak, and then declines to 58.9 mIoU. 
The best task trade-off is achieved with loss weights of 400, 80, and 200 for vehicle, drivable area, and lane, respectively; further increasing loss weights for vehicle and lane classes can hurt the segmentation performance of all three tasks.

\begin{figure}[!t]
    \centering
    \includegraphics[width=0.85\linewidth]{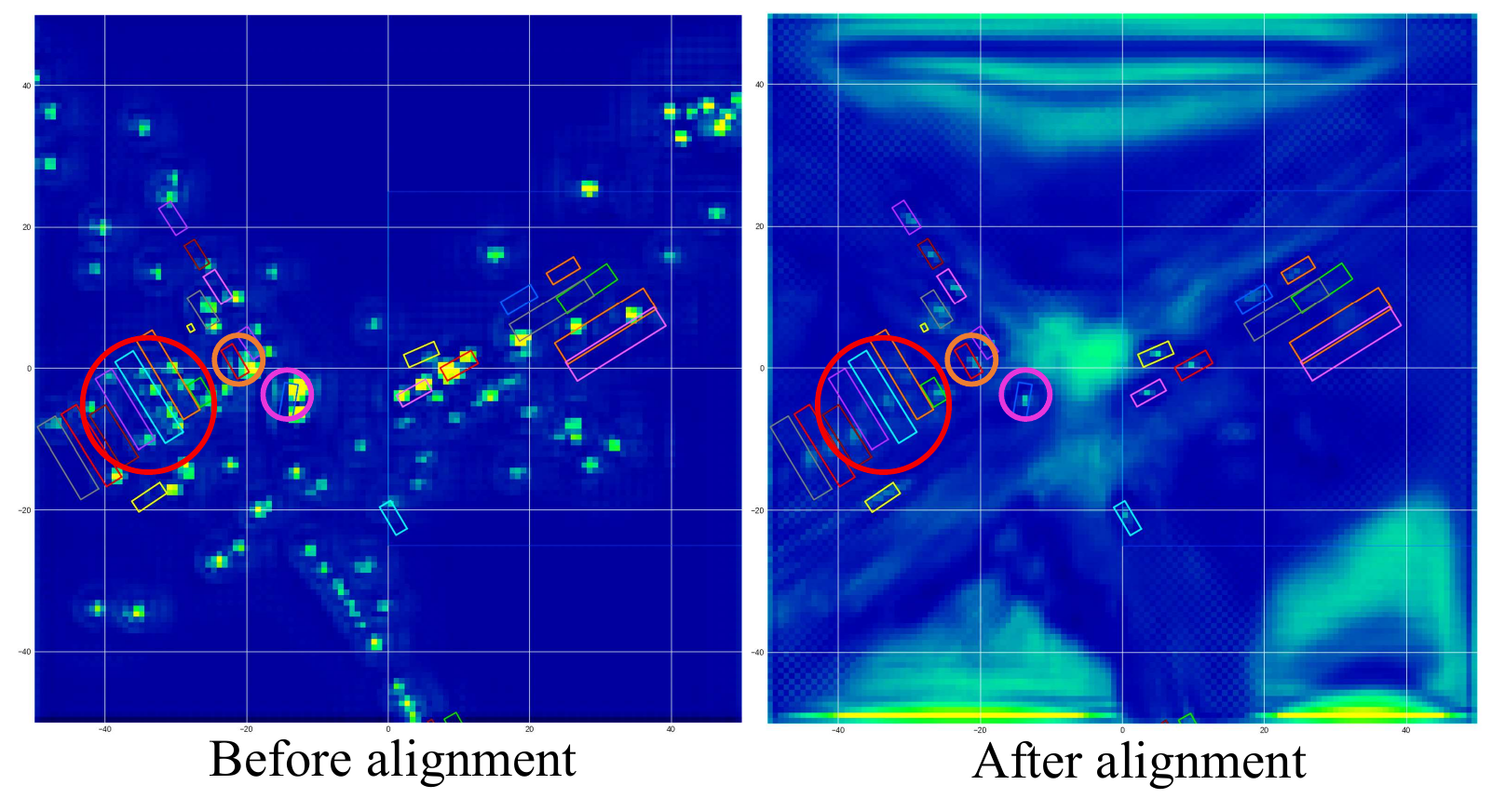}
    \caption{
    \textbf{Visualization of radar misalignment.} The boxes are the ground-truth boxes. CAMF can correct the radar misalignment.
    }
    \label{fig:mismatch}
\end{figure}

\renewcommand{\thefootnote}{1}
\setlength{\tabcolsep}{0.65em}
\begin{table*}[!t]
\begin{center}
\caption{
    \textbf{Task trade-off for BEV semantic segmentation}. The best task trade-off is obtained when the loss weights for the three tasks are 400, 80, and 200.
}
\label{table:seg trade-off}
\begin{tabular}{ccc||ccc|c}
    \hline
    \multicolumn{3}{c||}{Loss Weights} & \multicolumn{3}{c|}{IoU} & \multirow{2}{*}{mIoU$\uparrow$} \\
    Vehicle & Driv. Area & Lane & Vehicle$\uparrow$ & Driv. Area$\uparrow$ & Lane$\uparrow$ & \\
    \hline
    100 & 100 & 100 &  48.7 & 79.2 & 45.3 & 57.7\\  
    200 & 50 & 150 & 51.2 & 78.8 & \textbf{46.2} & 58.7\\ 
    400 & 80 & 200 & \textbf{52.8} & \textbf{79.8} & 46.0 & \textbf{59.5}\\ 
    500 & 75 & 175 & \textbf{52.8} & 78.7 & 44.6 & 58.7\\  
    600 & 70 & 300 & 52.7 & 78.1 & 45.9 & 58.9\\
    \hline

\end{tabular}
\end{center}
\end{table*}

\subsection{Analysis of Robustness}
\label{sec:robust}
\subsubsection{Sensor Failure}
To analyze the robustness in sensor failure scenarios, we randomly drop either images or radar inputs for evaluation.
%
In this experiment, we adopt the dropout training strategy as data augmentation to train RCBEVDet and report \textit{Car} class mAP following CRN~\cite{Kim_2023_ICCVCRN}.
Specifically, RCBEVDet consistently outperforms CRN and BEVFusion, showing higher mAP for the \textit{Car} class in all sensor failure cases.
Notably, the performance of CRN decreases by 4.5, 11.8, and 25.0 mAP in three radar sensor failure cases, while RCBEVDet only experiences drops of 0.9, 6.4, and 10.4 mAP.

These results highlight that the proposed cross-attention module enhances the robustness of BEV features through dynamic alignment.

\begin{table}[!t]
\caption{
\textbf{Analysis of robustness}. We report \textit{Car} class mAP in the same manner as CRN~\cite{Kim_2023_ICCVCRN}.}
\begin{center}
\label{table:robust}
\begin{tabular}{l|c|c||cccc}
\hline
& \multirow{2}{*}{Input} & \multirow{2}{*}{Drop} & \multicolumn{4}{c}{\# of view drops} \\
&  &  & 0 & 1 & 3 & All \\
\hline
BEVDepth~\cite{li2023bevdepth}    &  C  & C & 49.4 & 41.1 & 24.2 & 0 \\
CenterPoint~\cite{yin2021centerpoint} &  R  & R & 30.6 & 25.3 & 14.9 & 0 \\
\hline
\multirow{2}{*}{BEVFusion~\cite{DBLP:journals/corr/bevfusionmit}} & \multirow{2}{*}{C+R} & C & \multirow{2}{*}{63.9} 
& 58.5 & 45.7 & {14.3} \\
& & R & & 59.9 & 50.9 & 34.4 \\
\hline
\multirow{2}{*}{CRN~\cite{Kim_2023_ICCVCRN} }      & \multirow{2}{*}{C+R} & C & \multirow{2}{*}{\shortstack{{68.8}}} 
& {62.4} & {48.9} & 12.8 \\
& & R & & {64.3} & {57.0} & {43.8} \\
\hline
\multirow{2}{*}{\ours~(Ours)}       & \multirow{2}{*}{C+R} & C & \multirow{2}{*}{\shortstack{\textbf{72.5}}} 
& \textbf{66.9} & \textbf{53.5} & \textbf{16.5} \\

& & R & & \textbf{71.6} & \textbf{66.1} & \textbf{62.1} \\

\hline
\end{tabular}
\end{center}
\end{table}

\subsubsection{Modal Alignment}
To further demonstrate the effects of radar-camera alignment with CAMF, we randomly perturb the x- and y-axis coordinates of radar inputs.
Concretely, we uniformly sample noise from $[-1, 1]$ for the x-axis and y-axis coordinates of each radar point, respectively.
As shown in Table~\ref{table:align}, RCBEVDet only drops 1.3 NDS and 1.5 mAP with noise radar inputs, while CRN decreases by 2.3 NDS and 5.1 mAP. 
Besides, we visualize how CAMF addresses radar misalignment in Figure~\ref{fig:mismatch}.
As illustrated, many radar features exhibit positional offsets from ground-truth boxes. With CAMF, these radar features are realigned within the ground-truth boxes, effectively correcting the radar misalignment.

\begin{table}[!t]
\begin{center}
\caption{
    \textbf{Radar-camera multi-modal alignment with noise radar inputs.}
}
\label{table:align}
\begin{tabular}{c||c|c|c|c}
    \hline
    \multirow{2}{*}[-0.5ex]{Radar Input} & \multicolumn{2}{c|}{RCBEVDet}   & \multicolumn{2}{c}{CRN} \\
    \cline{2-5}
    & NDS$\uparrow$ & mAP$\uparrow$ & NDS$\uparrow$ & mAP$\uparrow$\\
    \hline
    Original & 56.8 & 45.3 & 56.0 & 49.0 \\
    Noise & 55.5 \textcolor{green}{$\downarrow$1.3} & 43.7 \textcolor{green}{$\downarrow$1.5} & 53.7 \textcolor{green}{$\downarrow$2.3} & 43.9 \textcolor{green}{$\downarrow$5.1} \\
    \hline
\end{tabular}
\end{center}
\end{table}

\subsubsection{Comparison with CRN}
\begin{figure}[!t]
    \centering
    \includegraphics[width=0.98\linewidth]{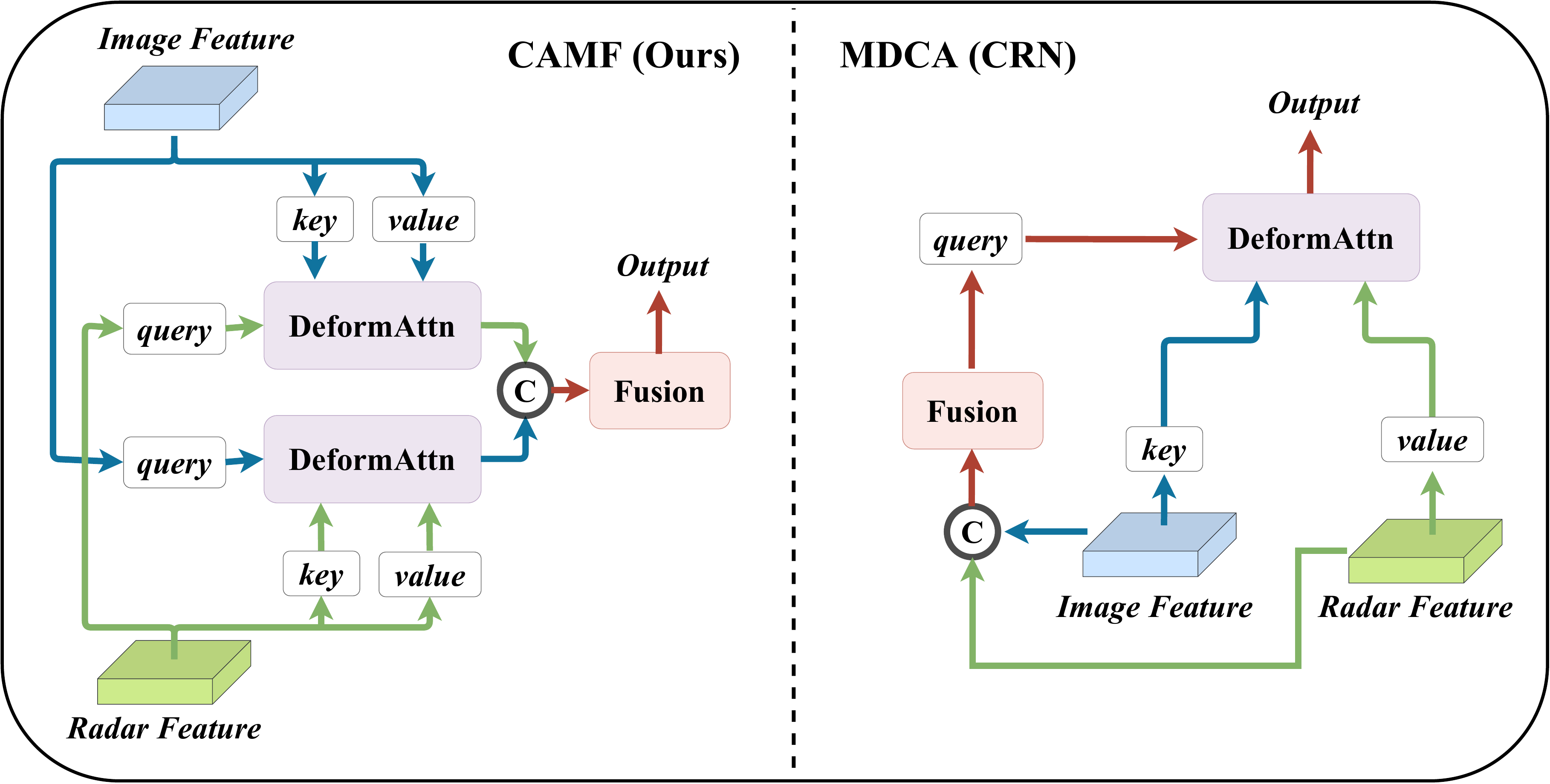}
    \caption{
    \textbf{Comparison between alignment modules in RCBEVDet and CRN.} RCBEVDet employs an alignment-then-fusion pipeline and obtains better robustness.
    }
    \label{fig:compare}
\end{figure}

CRN~\cite{Kim_2023_ICCVCRN} also utilizes deformable cross-attention to address the radar-camera mismatch issue.
The results in Tables \ref{table:robust} and~\ref{table:align} demonstrate that our CAMF is more robust than the Multi-modal Deformable cross-attention module (MDCA) proposed in CRN.
To further distinguish our method from CRN, we present the architecture details of the fusion module in Figure~\ref{fig:compare}.
Specifically, CAMF in RCBEVDet first aligns features from the camera and radar separately and then fuses aligned features using several convolutional blocks.
In contrast, MDCA in CRN first fuses features from two modalities. Then, it uses the fused features, image features, and radar features as the query, key, and value, respectively, and applies deformable attention to adjust the fused features.
Thus, the fusion-then-alignment pipeline of MDCA in CRN has limited alignment precision because it still has feature mismatches in the camera and radar fusion process.
In comparison, our CAMF, which employs an alignment-then-fusion pipeline, fully leverages the alignment capabilities of deformable attention in the first step and achieves better feature fusion robustness.

\subsection{Model Generalization}
\label{sec:generalization}
RCBEVDet employs a dual-branch architecture to fuse radar and multi-view cameras and can be incorporated with any existing multi-view camera-based methods to enhance feature representation.
To demonstrate the model generalization of RCBEVDet, we conduct experiments with various backbone and detector designs in the 3D object detection framework.

\subsubsection{Generalization for Backbone Architectures}
To demonstrate the RCBEVDet's model generalization for backbone architecture, we conduct experiments on BEVDepth with various backbone architectures, including CNN-based and Transformer-based backbones. 
As shown in Table \ref{table:backbone generalization}, our method can improve baseline performance by over 3.8$\sim$4.9 NDS and 4.8$\sim$10.2 mAP across different backbones.
Furthermore, for the same type of backbone architectures with varying sizes (\textit{e.g.}, ResNet-18 and ResNet-50), RCBEVDet can achieve consistent performance gains of 4.9 NDS.

\setlength{\tabcolsep}{0.65em}
\begin{table*}[!t]
\caption{
    \textbf{Model generalization capability of RCBEVDet over various special backbones.}
}
\begin{center}
\label{table:backbone generalization}
\begin{tabular}{l|c|c|c||cc}
    \hline
    Method & Input & Backbone & Image Size & NDS$\uparrow$ & mAP$\uparrow$ \\
    \hline 
    BEVDepth + Temporal
    & C & DLA34 & $448\times800$  & 52.5 & 40.2 \\
     \ours~(Ours) & C+R & DLA34 &  $448\times800$ & {56.3}\textcolor{red}{$\uparrow$3.8} & {45.3}\textcolor{red}{$\uparrow$5.1}  \\
    \hline
    BEVDepth + Temporal
    & C & Swin-T & $256\times704$ & 51.8 & 39.4 \\
     \ours~(Ours) & C+R & Swin-T &  $256\times704$ & {56.2}\textcolor{red}{$\uparrow$4.4}  & {49.6}\textcolor{red}{$\uparrow$10.2}  \\
    \hline
    BEVDepth + Temporal
    & C & ResNet-18 & $256\times704$ & 49.9& 35.0 \\
     \ours~(Ours) & C+R & ResNet-18 &  $256\times704$ &  {54.8}\textcolor{red}{$\uparrow$4.9}  & 42.9\textcolor{red}{$\uparrow$7.9}  \\
    \hline
    BEVDepth + Temporal
    & C & ResNet-50 & $256\times704$ & 51.9 & 40.5 \\
     \ours~(Ours) & C+R & ResNet-50 &  $256\times704$ & {56.8}\textcolor{red}{$\uparrow$4.9}  & 45.3\textcolor{red}{$\uparrow$4.8}  \\
    \hline
\end{tabular}
\end{center}
\end{table*}

\subsubsection{Generalization for 3D Detector Architecture}
We evaluate the detector architecture generalization of our method by plugging it into various mainstream multi-view camera-based 3D object detectors, including LSS-based (\textit{e.g.}, BEVDet and BEVDepth) and transformer-based methods (\textit{e.g.}, StreamPETR and SparseBEV).
These methods represent a range of detector designs.
As shown in Table \ref{table:head generalization}, our method boosts the performance of all popular multi-view camera-based 3D object detectors by fusing radar features.
Specifically, for the LSS-based method, RCBEVDet improves 5.6 NDS and 4.9 NDS for BEVDet and BEVDepth, respectively.
For the transformer-based detectors, RCBEVDet++ obtains similar performance gains in NDS, \textit{i.e.}, 5.6 NDS and 5.9 NDS improvement for StreamPETR and SparseBEV, respectively.
Notably, we observe more mAP improvements in transformer-based methods compared to LSS-based methods. 
%
The reason is that LSS-based methods typically use depth supervision from LiDAR points for more accurate 3D position prediction, while transformer-based methods learn 3D position implicitly.
Consequently, the transformer-based methods can derive more benefits from radar features, which provide additional depth information.
%
Overall, these results demonstrate the detector architecture generalization of our method across various 3D object detectors.

\setlength{\tabcolsep}{0.65em}
\begin{table}[!t]
\caption{
    \textbf{Model generalization capability over various multi-view camera-based 3D object detectors.} 
}
\begin{center}
\label{table:head generalization}
\begin{tabular}{l|c||cc}
    \hline
    Method & Input & NDS$\uparrow$ & mAP$\uparrow$ \\
    \hline 
    BEVDet + Temporal
    & C & 48.7 & 35.4 \\
     \ours~(Ours) & C+R &  54.3\textcolor{red}{$\uparrow$5.6}  & 41.3\textcolor{red}{$\uparrow$5.9}  \\
    \hline
    BEVDepth + Temporal
    & C & 51.9 & 40.5 \\
     \ours~(Ours) & C+R & 56.8\textcolor{red}{$\uparrow$4.9}  & 45.3\textcolor{red}{$\uparrow$4.8}  \\
    \hline
    StreamPETR & C & 54.0 & 43.2 \\
    RCBEVDet++~(Ours) & C+R & 59.6\textcolor{red}{$\uparrow$5.6}  & 51.5\textcolor{red}{$\uparrow$8.3}  \\
    \hline
    SparseBEV & C & 54.5 & 43.2 \\
    RCBEVDet++~(Ours) & C+R & 60.4\textcolor{red}{$\uparrow$5.9}  & 51.9\textcolor{red}{$\uparrow$8.7}  \\
    \hline
\end{tabular}
\end{center}
\end{table}

\section{Conclusion}
In this paper, we first introduce RCBEVDet, a radar-camera fusion 3D detector.
It comprises an existing camera-based 3D detection model, a specially designed radar feature extractor, and the CAMF module for aligning and fusing radar-camera multi-modal features.
RCBEVDet improves the performance of various camera-based 3D object detectors across several backbones and shows well robustness capability against sensor failure cases on the nuScenes dataset.

To unleash the full potential of RCBEVDet, we propose RCBEVDet++, which extends the CAMF module to support query-based multi-view camera perception models with sparse fusion and adapts to more perception tasks, including 3D object detection, BEV semantic segmentation, and 3D multi-object tracking.
Extensive experiments on the nuScenes dataset show that RCBEVDet++ further boosts the performance of camera-based perception models and achieves new state-of-the-art radar-camera multi-modal results on these three perception tasks.
Notably, without using test-time augmentation or model ensemble, RCBEVDet++ obtains 72.73 NDS and 67.34 mAP for 3D object detection with ViT-L as the image backbone.


%





\ifCLASSOPTIONcaptionsoff
  \newpage
\fi



%



\bibliographystyle{IEEEtran}
\bibliography{IEEEabrv,main}

%








\end{document}